\documentclass{article}

    \usepackage[preprint]{neurips_2025}

\usepackage[utf8]{inputenc} % allow utf-8 input
\usepackage[T1]{fontenc}    % use 8-bit T1 fonts
\usepackage[unicode]{hyperref}       % hyperlinks
\usepackage{url}            % simple URL typesetting
\usepackage{amsfonts}       % blackboard math symbols
\usepackage{nicefrac}       % compact symbols for 1/2, etc.
\usepackage{microtype}      % microtypography
\usepackage{graphicx}
\usepackage{multirow}
\usepackage{booktabs}       % professional-quality tables
\usepackage[normalem]{ulem}
\useunder{\uline}{\ul}{}
\usepackage[table,xcdraw]{xcolor}
\newcolumntype{P}[1]{>{\hspace{#1}}c<{\hspace{#1}}}  % centered

\usepackage{caption}

\usepackage{amsmath}
\usepackage{amssymb}
\usepackage{mathtools}
\usepackage{amsthm}
\usepackage{bbm}
\DeclareMathOperator*{\argmin}{arg\,min}
\usepackage[mathscr]{euscript}
\usepackage{enumitem}
\usepackage{dblfloatfix} % for putting table at bottom
\newcommand{\indep}{\perp \!\!\! \perp}

\theoremstyle{plain}
\newtheorem{theorem}{Theorem}[section]

\newtheorem{corollary}[theorem]{Corollary}
\theoremstyle{definition}

\theoremstyle{remark}

\usepackage[normalem]{ulem}

\usepackage{algorithm}      % defines the floating “algorithm” environment
\usepackage{algorithmic}    % defines the “algorithmic” (pseudo-code) environment

\title{RDIT: Residual-based Diffusion Implicit Models for Probabilistic Time Series Forecasting}

\author{%
    Chih-Yu Lai$^*$ \\
    Dept. of EECS, MIT \\
    Cambridge, MA, 02139 \\
    \texttt{chihyul@mit.edu} \\
  \And
    Yu-Chien Ning \\
    Harvard University \\
    Cambridge, MA, 02138 \\
    \texttt{bycning@hsph.harvard.edu} \\
  \And
    Duane S. Boning \\
    Dept. of EECS, MIT \\
    Cambridge, MA, 02139 \\
    \texttt{boning@mtl.mit.edu}
}

\begin{document}

\maketitle

\begin{abstract}
Probabilistic Time Series Forecasting (PTSF) plays a critical role in domains requiring accurate and uncertainty-aware predictions for decision-making. However, existing methods offer suboptimal distribution modeling and suffer from a mismatch between training and evaluation metrics. Surprisingly, we found that augmenting a strong point estimator with a zero-mean Gaussian, whose standard deviation matches its training error, can yield state-of-the-art performance in PTSF. In this work, we propose RDIT, a plug-and-play framework that combines point estimation and residual-based conditional diffusion with a bidirectional Mamba network. We theoretically prove that the Continuous Ranked Probability Score (CRPS) can be minimized by adjusting to an optimal standard deviation and then derive algorithms to achieve distribution matching. Evaluations on eight multivariate datasets across varied forecasting horizons demonstrate that RDIT achieves lower CRPS, rapid inference, and improved coverage compared to strong baselines. (Code can be downloaded at \href{https://anonymous.4open.science/r/RDIT-16BB/}{https://anonymous.4open.science/r/RDIT-16BB/})
\end{abstract}

%%%%%%%%%%%%%%%%%%%%%%%%%%%%%%%%%%%%%%%%%%%%%%%%%%%%%%%%%%%%%%%%%%%%%%%%%%%%%%%%%%
%%%%%%%%%%%%%%%%%%%%%%%%%%%%%%%%%%%%%%%%%%%%%%%%%%%%%%%%%%%%%%%%%%%%%%%%%%%%%%%%%%
%%%%%%%%%%%%%%%%%%%%%%%%%%%%%%%%%%%%%%%%%%%%%%%%%%%%%%%%%%%%%%%%%%%%%%%%%%%%%%%%%%
\section{Introduction}
\label{sec:intro}

Time Series Forecasting (TSF) aims to predict future values of variables based on their historical observations. This challenging but important task can be widely applied to a variety of fields such as finance \cite{SEZER2020106181}, healthcare \cite{10.1145/3531326}, manufacturing \cite{machines12060380}, environmental science \cite{Brbulescu2016StudiesOT}, and many others \cite{10053870,Karl2024,mao2024time}. An emerging amount of literature focuses on implementing deep learning-based techniques for TSF \cite{miller2024survey,doi:10.1098/rsta.2020.0209,doi:10.1089/big.2020.0159}. PTSF can be applied to additionally model the distribution of the predictions for estimating the uncertainty \cite{probforecastreview}, where several advantages include increased robustness, the ability of risk management, and the likelihood of anomaly detection \cite{electricity2010002,tyralis2022review}. Diffusion-based generative models are a common approach for realizing PTSF \cite{Lin2024,kollovieh2023predict}. This involves iteratively denoising the data and arriving at some predictive pattern, where the end points of multiple trajectories form a distribution of the predictions. These models exhibit strong distribution modeling characteristics and have achieved state-of-the-art performance \cite{meijer2024rise}.

However, certain techniques shared among recent TSF methods inherently pose challenges for distributional modeling in PTSF~\cite{PatchTST,li2023revisiting,toner2024analysis}. Specifically, linear mappings may fail to capture nonlinear relationships or interactions that are critical for accurately modeling the distribution of the forecast~\cite{li2023revisiting}, and channel independence can prevent the model from learning dependencies among variables, resulting in unrealistic uncertainty estimates~\cite{rasul2020multivariate}. While recent TSF methods do not strictly enforce the use of these mechanisms~\cite{SMamba,iTransformer,shang2024ada}, integrating both point estimation and noise estimation within a single model often leads to a trade-off between prediction accuracy and diversity, making it difficult to model uncertainty effectively~\cite{TMDM}. To address this issue, TMDM \cite{TMDM} separates the tasks of point-based (conditional mean) estimation and noise estimation: the former is predicted using a plug-and-play transformer-based model, and the latter is modeled using diffusion. This separation allows any TSF model to be integrated for point estimation while enabling more targeted design of the noise estimation process. Recently, D3U~\cite{D3U} improves upon TMDM by incorporating a better-performing point estimator, residual modeling, and explicitly decoupling deterministic and uncertain components. It also adopts the DPM-Solver~\cite{lu2022dpm} for faster inference.

Another challenge is that as the CRPS is a widely used metric for evaluating PTSF, using the Mean Absolute Error (MAE) or Mean Squared Error (MSE) as the training loss can lead to a mismatch between the training objective and the evaluation metric. This misalignment may result in under-dispersed or overconfident predictions~\cite{zheng2024mvg}, suggesting that post-hoc adjustments to the noise estimates are necessary to optimize the final predictive distribution. Additionally, since the residuals contain inherent noise, it is easy for models to overfit to them, leading to a potential mismatch between the Predictive Interval Coverage Probability (PICP) of the predicted and true distributions. These challenges can result in suboptimal performance. What's more, we find that a strong baseline can be constructed by equipping a point estimator with a zero-mean Gaussian distribution fitted to the training residuals, i.e., $\mathcal{N}(0, \boldsymbol\sigma_{trn}^2)$, which achieves excellent results in several settings (Section~\ref{ssec:main}). This introduces a \textit{race} between improving point estimation and uncertainty modeling. A robust PTSF algorithm should outperform its corresponding point estimator in both point-based and probabilistic metrics even when the point estimator is enhanced with a fitted Gaussian.

In this work, we propose \textbf{R}esidual-based \textbf{D}iffusion \textbf{I}mplicit Models for Probabilistic \textbf{T}ime Series Forecasting (RDIT), a framework that leverages point estimators for accurate point forecasting and noise estimation networks to model the residuals between the ground truth and point predictions. To address the distribution mismatch problem, we introduce two novel algorithms: 
1) Error‐aware Expansion (EAE), which mitigates the discrepancy between MAE and CRPS by optimizing the predicted variance, and 2) Coverage Optimization (CO), which calibrates on a hold-out validation set to adjust the final distribution. RDIT incorporates Denoising Diffusion Implicit Models (DDIMs) \cite{DDIM} for fast inference and introduces the PICP distance metric, which measures the total deviation from the target coverage across multiple quantiles (cf. Appendix~\ref{sec:metrics}). Our experiments demonstrate that RDIT outperforms ten baselines across eight datasets and multiple prediction lengths. 

{\bf Key contributions} of this paper are:
\begin{enumerate}
    \item We decouple point estimation and residual modeling into two stages---using TSF models to generate point estimates and applying diffusion to normalized residuals---to improve learnability by minimizing the discrepancy between the start and end of the diffusion process.
    % \ycncmt{I removed ``recent'' before ``TSF models'' to save space by reducing one line.}
    \item We propose two distribution-matching algorithms: (a) EAE that minimizes the CRPS under a fixed-mean Gaussian predictive distribution by optimizing the variance, and (b) CO, a coverage optimization algorithm that corrects mismatches between true and predictive coverage. Both contribute to improved performance.
    \item We design a noise estimation network using bidirectional Mamba layers to effectively model temporal dependencies in the residuals, leveraging correlations between the denoising target and the input to enhance inductive bias.
    % this network is specifically designed for noise estimation, so i think this is a novel point
    % \ycncmt{Is this new, or has been already done in the Mamba paper by Gu \& Dao (2024) already?}
\end{enumerate}

%%%%%%%%%%%%%%%%%%%%%%%%%%%%%%%%%%%%%%%%%%%%%%%%%%%%%%%%%%%%%%%%%%%%%%%%%%%%%%%%%%
%%%%%%%%%%%%%%%%%%%%%%%%%%%%%%%%%%%%%%%%%%%%%%%%%%%%%%%%%%%%%%%%%%%%%%%%%%%%%%%%%%
%%%%%%%%%%%%%%%%%%%%%%%%%%%%%%%%%%%%%%%%%%%%%%%%%%%%%%%%%%%%%%%%%%%%%%%%%%%%%%%%%%
%\ycncmt{on Denoising Diffusion Models}
\section{Preliminaries}
\label{sec:prelim}

\subsection{Denoising Diffusion Probabilistic Models (DDPMs)}
\label{ssec:DDPM}

% checked grammar, is correct
DDPMs define a forward diffusion process where an initial data point $\mathbf{r}^0$ is gradually corrupted over $K$ diffusion steps by adding Gaussian noise. At each step $k \in \{1,2,\dots,K\}$, the data is transformed according to $q(\mathbf{r}^k | \mathbf{r}^{k-1}) = \mathcal{N}\left(\sqrt{1-\beta_k}\mathbf{r}^{k-1}, \beta_k \mathbf{I}\right)$, where $\beta_k \in [0,1]$ is a variance schedule parameter controlling the amount of noise added step $k$. By leveraging the properties of the Gaussian distribution and the variational approximation, the marginal distribution of $\mathbf{r}^k$ given $\mathbf{r}^0$ is
\begin{align}
\label{eq:forward_sample_k}
    q(\mathbf{r}^k | \mathbf{r}^{0}) = \mathcal{N}(\sqrt{\Bar{\alpha}_k}\mathbf{r}^{0}, (1-\Bar{\alpha}_k) \mathbf{I}), \; \mathrm{with} \;
    \Bar{\alpha}_k = \prod_{s=1}^{k} \alpha_{s}, \; \alpha_s = 1 - \beta_s.
\end{align}
Thus, $\mathbf{r}^k$ can be directly sampled from $\mathbf{r}^0$ using:
\begin{gather}
\label{eq:rk_sample}
    \mathbf{r}^k = \sqrt{\bar{\alpha}_k} \mathbf{r}^0 + \sqrt{1 - \bar{\alpha}_k} \boldsymbol{\varepsilon}^k, \quad \boldsymbol{\varepsilon}^k \sim \mathcal{N}(0, \mathbf{I}).
\end{gather}
The reverse process aims to recover the original data $\mathbf{r}^0$ from pure noise $\mathbf{r}^K$, where $\mathbf{r}^K$ is sampled from the standard Gaussian distribution $\mathcal{N}(0, \mathbf{I})$. It is modeled as a Markov chain with learned Gaussian transitions:
$p_{\theta}(\mathbf{r}^{k-1} | \mathbf{r}^k) = \mathcal{N}(\boldsymbol{\mu}_{\theta}(\mathbf{r}^k, k), \sigma_k^2 \mathbf{I})$, where $\sigma_k^2$ is the variance, and $\boldsymbol{\mu}_\theta(\mathbf{r}^k, k)$ is calculated from the output of a neural network parameterized by $\theta$ ($\boldsymbol{\varepsilon}_\theta$).
In the noise estimation framework~\cite{DDPM}, the mean is computed as
\begin{gather}
\label{eq:reverse_mean_noise_estimation}
    \boldsymbol{\mu}_\theta(\mathbf{r}^k, k) = \frac{1}{\sqrt{\alpha_k}} \left( \mathbf{r}^k - \frac{\beta_k}{\sqrt{1 - \bar{\alpha}_k}} \boldsymbol{\varepsilon}_\theta(\mathbf{r}^k, k) \right),
\end{gather}
where $\boldsymbol{\varepsilon}_\theta(\mathbf{r}^k, k)$ predicts the added noise at step $k$.
The parameters $\theta$ can be learned by minimizing the MAE (denoted by $\mathscr{L}(\theta)$) between the true and predicted values:
\begin{gather}
\label{eq:DDPM_noise_loss}
    \mathscr{L}(\theta) = \mathbb{E}_{k, \mathbf{r}^0, \boldsymbol{\varepsilon}} \left[\| \boldsymbol{\varepsilon}^k - \boldsymbol{\varepsilon}_\theta(\mathbf{r}^k, k)\|_1 \right].
\end{gather}
Alternatively, in the data prediction framework, we have:
$\boldsymbol{\mu}_\theta(\mathbf{r}^k, k) = \sqrt{\alpha_k} (1 - \bar{\alpha}_{k-1})\mathbf{r}^k/(1 - \bar{\alpha}_k) + 
\sqrt{\bar{\alpha}_{k-1}} \beta_k \mathbf{r}_\theta(\mathbf{r}^k, k)/(1 - \bar{\alpha}_k)$,
where $\mathbf{r}_\theta(\mathbf{r}^k, k)$ is a network that directly estimates $\mathbf{r}^0$ (see \cite{DDPM}).

\subsection{Denoising Diffusion Implicit Models (DDIMs)}
\label{ssec:DDIM}
Rather than formulating the inference process as a Markov process, DDIMs \cite{DDIM} define a non-Markovian inference process that leads to the same surrogate objective function as DDPMs: $q_\sigma(\mathbf{r}^{1:K}|\mathbf{r}^0) = q_\sigma(\mathbf{r}^K|\mathbf{r}^0) \prod_{k=2}^{K} q_\sigma(\mathbf{r}^{k-1}|\mathbf{r}^k, \mathbf{r}^0)$, where
$q_\sigma(\mathbf{r}^K | \mathbf{r}^{0}) = \mathcal{N}(\sqrt{\Bar{\alpha}_K}\mathbf{r}^{0}, (1-\Bar{\alpha}_K) \mathbf{I})$,
and for all $k > 1$,
\begin{gather}
\label{eq:rkminusone-cond-rk}
q_\sigma(\mathbf{r}^{k-1}|\mathbf{r}^k, \mathbf{r}^0) =  
    \mathcal{N}\left(\sqrt{\Bar{\alpha}_{k-1}}\mathbf{r}^0 
    + \sqrt{\frac{1 - \Bar{\alpha}_{k-1} - \sigma^2_k}{1 - \Bar{\alpha}_k}} \left(\mathbf{r}^k - \sqrt{\Bar{\alpha}_k} \mathbf{r}^0\right), \ \sigma_k^2 \mathbf{I}\right).
\end{gather}
% \ycncmt{Is $1 - \bar \alpha_{k-1} - \sigma_k^2$ guaranteed to be positive?}
The mean is chosen such that (\ref{eq:forward_sample_k}) is satisfied for all $k > 1$. In our experiments, $\sigma_k$ is set to 0, and uncertainty arises purely from the diffusion endpoint $\mathbf{r}^K$ during training and inference. To accelerate the generation process, the inference process is factored as the following non-Markovian process:
$q_{\sigma,\kappa}(\mathbf{r}^{1:K}|\mathbf{r}^0) = 
    q_{\sigma,\kappa}(\mathbf{r}^{\kappa_W}|\mathbf{r}^0) \prod_{i=1}^{W} q_{\sigma,\kappa}(\mathbf{r}^{\kappa_{i-1}}|\mathbf{r}^{\kappa_i}, \mathbf{r}^0) \prod_{k \in \Bar{\kappa}} q_{\sigma,\kappa}(\mathbf{r}^k|\mathbf{r}^0)$,
where $\kappa$ is a sub-sequence of $1, \dots, K$ with length $W \leq K$ and $\kappa_W = K$, and $\Bar{\kappa} = \{1, \dots, K\} \backslash \kappa$. 
With the above process, we define
\begin{gather}
    q_{\sigma,\kappa}(\mathbf{r}^k|\mathbf{r}^0) = \mathcal{N}(\sqrt{\Bar{\alpha}_k}\mathbf{r}^{0}, (1-\Bar{\alpha}_k) \mathbf{I}) \;\; \forall k \in \Bar{\kappa} \cup \{K\},
\label{eq:acc_sample_not_on_traj}
\end{gather}
% already defined above (length of \kappa)
% \ycncmt{need to define $[W]$ check!}
and for all $i \in \{1, \dots, W\}$, the probability of $\mathbf{r}^{\kappa_{i-1}}$ conditioned on $\mathbf{r}^{\kappa_i}$ and $\mathbf{r}^0$ is
\begin{equation*}
    q_{\sigma,\kappa}(\mathbf{r}^{\kappa_{i-1}}|\mathbf{r}^{\kappa_i}, \mathbf{r}^0) = \mathcal{N}\left(\sqrt{\Bar{\alpha}_{\kappa_{i-1}}}\mathbf{r}^0
    + \sqrt{\frac{1 - \Bar{\alpha}_{\kappa_{i-1}} - \sigma^2_{\kappa_i}}{1 - \Bar{\alpha}_{\kappa_i}}} 
    \left(\mathbf{r}^{\kappa_i} - \sqrt{\Bar{\alpha}_{\kappa_i}} \mathbf{r}^0\right),\  \sigma_{\kappa_i}^2 \mathbf{I}\right),
\end{equation*}
where $\kappa_0 = 0$, $\sigma_{k_i}$ is again set to 0, and the coefficients are chosen such that
\begin{gather}
    q_{\sigma,\kappa}(\mathbf{r}^{\kappa_i} | \mathbf{r}^0) = \mathcal{N}(\sqrt{\Bar{\alpha}_{\kappa_i}} \mathbf{r}^0, (1 - \Bar{\alpha}_{\kappa_i}) \mathbf{I}) \;\; \forall i \in \{1, \dots, W\}.
\label{eq:acc_sample_on_traj}
\end{gather}
% \ycncmt{I am unsure how to get (2), need to be more specific.}.
Combining (\ref{eq:acc_sample_not_on_traj}) and (\ref{eq:acc_sample_on_traj}), we have (\ref{eq:forward_sample_k}) (refer to \cite{DDIM} for a more detailed derivation). Hence, the training objective can be the same as in DDPMs. During inference, $\kappa$ can be selected to best balance between generation efficiency and diversity. Due to conflicts of naming conventions between TSF and DDPM, we use $k, K, \mathbf{r}$ rather than $t, T, \mathbf{x}$ in our definitions.

%%%%%%%%%%%%%%%%%%%%%%%%%%%%%%%%%%%%%%%%%%%%%%%%%%%%%%%%%%%%%%%%%%%%%%%%%%%%%%%%%%
%%%%%%%%%%%%%%%%%%%%%%%%%%%%%%%%%%%%%%%%%%%%%%%%%%%%%%%%%%%%%%%%%%%%%%%%%%%%%%%%%%
%%%%%%%%%%%%%%%%%%%%%%%%%%%%%%%%%%%%%%%%%%%%%%%%%%%%%%%%%%%%%%%%%%%%%%%%%%%%%%%%%%

\section{RDIT: Residual Diffusion Implicit Model with Distribution Matching}
\label{sec:method}

\begin{figure*}
    \centering
    \includegraphics[width=\linewidth]{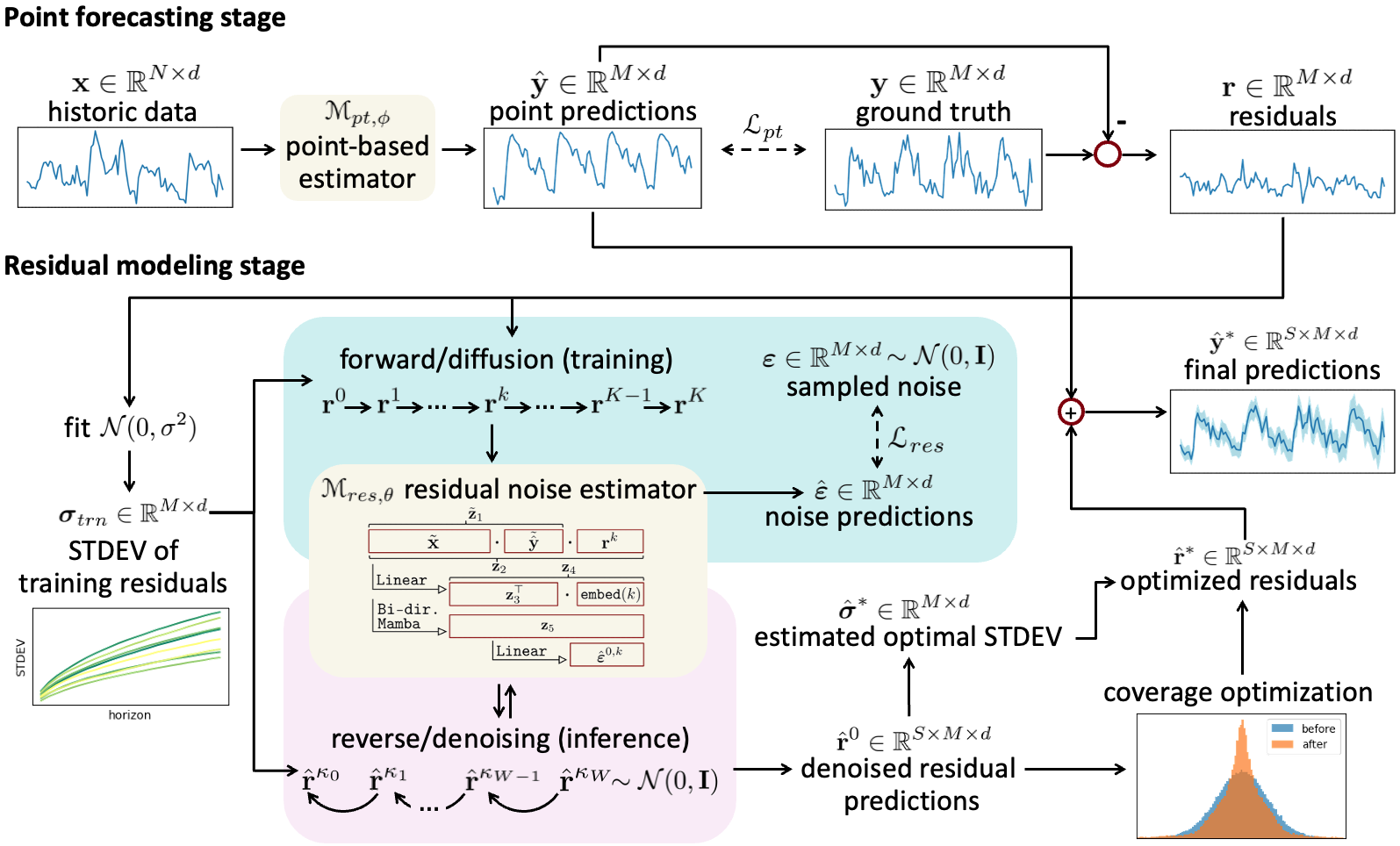}
    \caption{General scheme of this work. The point estimator ($\mathscr{M}_{pt,\phi}$) is used for predicting the point estimates ($\hat{\mathbf{y}}$), and the residual-based conditional model ($\mathscr{M}_{res,\theta}$) is used for predicting the noise ($\hat{\boldsymbol{\varepsilon}}$) from the $k$-th diffusion step residual ($\mathbf{r}^k$) while conditioning on $\mathbf{x}$ and $\hat{\mathbf{y}}$. Two loss functions $\mathscr{L}_{pt}$ and $\mathscr{L}_{res}$ are optimized for training the models. During inference, $\hat{\mathbf{r}}^{\kappa_W}$ is sampled from $\mathcal{N}(0, \mathbf{I})$, and the DDIM framework is applied for accelerated denoising. The standard deviation of the residuals ($\boldsymbol{\sigma}_{trn}$) is used for normalization of the residuals. The estimated optimal standard deviation ($\hat{\boldsymbol{\sigma}}^*$) is calculated from $\hat{\mathbf{r}}^0$ and is used to adjust the standard deviation after coverage optimization to yield the final predictions ($\hat{\mathbf{y}}^*$).}
    % done
    % \ycncmt{The notation $N(0, \sigma)$ in the figure should be $N(0, \sigma^2)$ in order to be consistent with the notations in Section 2.}
    \label{fig:scheme}
\end{figure*}

RDIT is a novel PTSF framework that incorporates distribution matching to optimize uncertainty estimates. This framework employs a two-stage procedure in both training and inference, as shown in Figure ~\ref{fig:scheme}. In the first stage, a point‐based estimator ($\mathscr{M}_{pt,\phi}$) produces point predictions of the future time series ($\hat{\mathbf{y}}$). In the second stage, a residual‐conditional model ($\mathscr{M}_{res,\theta}$) (analogous to $\boldsymbol{\varepsilon}_{\theta}$ in (\ref{eq:reverse_mean_noise_estimation})), models the residuals between the ground truth ($\mathbf{y}$) and $\hat{\mathbf{y}}$.

We begin by describing how $\mathscr{M}_{pt,\phi}$ is used to compute essential quantities in Section \ref{ssec:pb-estimator}, such as the residuals ($\mathbf{r}$), and the standard deviation of training errors ($\boldsymbol{\sigma}_{trn}$). Next, we explain how the residual‐based diffusion process models the distribution of residuals using these quantities, and how accelerated inference is performed in Section \ref{ssec:resdiff}. Afterward, we introduce the two distribution‐matching techniques, EAE and CO, to refine the prediction distribution in Section \ref{ssec:dist_match}. We then detail the neural network architecture of $\mathscr{M}_{res,\theta}$ in Section \ref{ssec:arch}. Finally, we summarize the overall training and inference workflow and provide pseudocode in Section \ref{ssec:train_test}; further details about the problem statement of PTSF is in Appendix \ref{sec:PTSF}.

\subsection{Point-based Estimator}
\label{ssec:pb-estimator}
%\ycncmt{Avoid starting a sentence with a mathematical symbol. Add an introductory sentence that outlines the theme of this subsection.}
The point estimator $\mathscr{M}_{pt,\phi}$ provides an initial point estimate for $\mathbf{y}$: $\hat{\mathbf{y}} = \mathscr{M}_{pt,\phi}(\mathbf{x})$, where the predictions are optimized by minimizing the MAE (Appendix \ref{sec:metrics}) between $\mathbf{y}$ and $\hat{\mathbf{y}}$, leading to an estimator for the conditional median \cite{Soch2025-ye}. Two important quantities can be calculated using $\mathscr{M}_{pt,\phi}$: the ground truth residuals ($\mathbf{r}$) and standard deviation of the training residuals ($\boldsymbol{\sigma}_{trn}$). Given point estimations $\hat{\mathbf{y}}$, we calculate $\mathbf{r}$ as $\mathbf{r} = \mathbf{y} - \hat{\mathbf{y}}$ and $\boldsymbol{\sigma}_{trn}$ by modeling $\mathbf{r}$ with a zero-mean Gaussian distribution using the training dataset. We obtain the standard deviation for each variate and time point in the prediction horizon, hence $\boldsymbol{\sigma}_{trn} \in \mathbb{R}^{M \times d}$. Generally, $\boldsymbol{\sigma}_{trn}$ gets larger as the time point gets farther. This holds if the dataset lacks strong periodicity and becomes more unpredictable over time (c.f. Figure \ref{fig:exchange_pred_itvl}). However, if the periodicity is strong, the prediction uncertainty might almost stay the same (c.f. Figure \ref{fig:ETTh1_pred_itvl}). Therefore, $\boldsymbol{\sigma}_{trn}$ is used to normalize the quantities during residual-based diffusion for more consistent distributions across time and variates (Section \ref{ssec:resdiff}). Using $\mathscr{M}_{pt,\phi}$ has three advantages: First, we can leverage recent TSF models as prior knowledge for estimating the conditional median \cite{TMDM,D3U}. Second, the other model, $\mathscr{M}_{res,\theta}$, can be specifically designed to capture the distribution of $\mathbf{r}$, alleviating the burden of both point estimation and distributional modeling. Last, focusing solely on the residuals enables the application of distribution matching (Section \ref{ssec:dist_match}) to enhance performance.

% \ycncmt{missing the link between using the point estimator and the RDIT approach}

\subsection{Residual-based Implicit Diffusion}
\label{ssec:resdiff}

In the second stage of our framework, we aim to model the distribution of $\mathbf{r}$ using conditional diffusion. The starting point of diffusion is the normalized residual: $\mathbf{r}^0 = \mathbf{r} / \boldsymbol{\sigma}_{trn}$. Our approach is similar to the noise prediction framework defined in (\ref{eq:reverse_mean_noise_estimation}), but additionally conditioning on the historic data ($\mathbf{x}$) and $\hat{\mathbf{y}}$:
\begin{gather}
\label{eq:cond_noise_pred}
    \hat{\boldsymbol{\varepsilon}}^{0,k} = \mathscr{M}_{res,\theta}(\mathbf{r}^k, k, \mathbf{x}, \hat{\mathbf{y}}),
\end{gather}
where $\mathbf{r}^k$ is sampled as in (\ref{eq:rk_sample}) and the loss is defined as in (\ref{eq:DDPM_noise_loss}).

In addition to the distribution matching algorithms proposed in Section \ref{ssec:dist_match}, we improve upon the residual diffusion framework proposed in D3U \cite{D3U} by using the \textit{normalized} residuals, $\mathbf{r}^0$, rather than raw residuals, as both the diffusion start point and denoising target. This ensures that both endpoints of the diffusion process share identical statistics, simplifying training and stabilizing the model.

% \ycncmt{DDIMs}
To address the major disadvantage of DDPMs (Section \ref{ssec:DDPM}), where significant slow-down occurs as inference time scales linearly with the number of diffusion steps $K$, we adopt the DDIM framework for accelerated inference (Section~\ref{ssec:DDIM}) and employ a subsequence $\kappa$ with reduced length $W < K$ to balance generation diversity with inference speed. $\sigma_{\kappa_i}$ in \eqref{eq:rkminusone-cond-rk} is set to 0 $\forall i \in \{1,2,\cdots,W\}$, and the corresponding denoising step from $\kappa_i$ to $\kappa_{i-1}$ is defined as
\begin{gather}
\label{eq:r_skip_pred}
    \hat{\mathbf{r}}^{\kappa_{i-1}} = \sqrt{\Bar{\alpha}_{\kappa_{i-1}}}\hat{\mathbf{r}}^{0,\kappa_i} + 
    \sqrt{
    \frac{1 - \Bar{\alpha}_{\kappa_{i-1}}}{1 - \Bar{\alpha}_{\kappa_i}}
    }
    \left(\hat{\mathbf{r}}^{\kappa_i} - \sqrt{\Bar{\alpha}_{\kappa_i}} \hat{\mathbf{r}}^{0,\kappa_i}\right),
\end{gather}
where $\hat{\mathbf{r}}^{\kappa_{i-1}}$ is the prediction at step $k = \kappa_{i-1}$, $\hat{\mathbf{r}}^{\kappa_0} = \hat{\mathbf{r}}^{0} = \hat{\mathbf{r}}/\boldsymbol\sigma_{trn}$ is the generation endpoint, $\hat{\mathbf{r}}^{0,\kappa_i}$ is the normalized predicted ground truth using (\ref{eq:reverse_mean_noise_estimation}) and (\ref{eq:cond_noise_pred}) with $k = \kappa_i$, and $\Bar{\alpha}_{\kappa_i}$ is calculated using (\ref{eq:forward_sample_k}) with $k = \kappa_i$. Since the denoising endpoint $\hat{\mathbf{r}}^0$ is already normalized, we constrain the intermediate outcomes to have zero mean and unit variance; thus, only the shape of the distribution changes during denoising. While increasing the number of diffusion steps $\kappa$ can improve sample fidelity, too many steps may cause target deviation or artifacts, leading to a more erroneous mean. In our experiments, a $\kappa$ with $W \approx 10$ suffices for effective diffusion, as $\mathscr{M}_{pt,\phi}$ already provides a reliable median estimate. Additional experiments show that, in general, the denoising process gradually improves the CRPS at each denoising step (Appendix~\ref{sec:CRPS_vs_denoise_step}).

% fixed penultimate paragraph so doesn't have repeated statements
% \ycncmt{Doesn't this paragraph repeat Point (2) in the penultimate paragraph?}

\subsection{Distribution Matching}
\label{ssec:dist_match}

% In this section, we highlight two potential drawbacks that may lead to suboptimal performance in current PTSF methods and propose two corresponding algorithms to address them: Error‐aware Expansion (EAE), which resolves the mismatch between training and evaluation metrics; and Coverage Optimization (CO), which adjusts miscalibrated prediction intervals arising from modeling bias.

In this section, we introduce the EAE and CO algorithms to address potential drawbacks that can lead to suboptimal performance in current PTSF methods.

%%%%%%%%%%%%%%%%%%%%%%%%%%%%%%%%%%%%%%%%%%
\subsubsection{Error-aware Expansion (EAE)}
\label{sssec:err_aware_exp}

An important metric used for evaluating the performance of probabilistic forecasts is the CRPS \cite{CRPS} (Appendix \ref{sec:metrics}). During training, minimizing MAE optimizes the accuracy of point estimates, but it disregards the uncertainty or spread of the predictions. This can lead to overconfident predictions. The CRPS, on the other hand, rewards models that provide a well-calibrated probabilistic distribution. A model that minimizes MAE might produce predictions that are tightly clustered around the median or mean, potentially leading to poorer probabilistic forecasts when evaluated with CRPS. One method to mitigate this issue is to directly minimize the CRPS during training \cite{zheng2024mvg}. However, multiple values for the same ground truth must be sampled in order to accurately calculate the CRPS, resulting in slower training. We thus introduce the EAE method to modify the original predictions and match the optimal distribution that minimizes the CRPS. This becomes feasible given the following theorem:
\begin{theorem}
\label{thm:best_sigma_CRPS}
Given ground truth $y$ and predictions $\hat y$, where $\hat y \sim \mathcal{N}(0, \sigma^2)$, where $\sigma$ is the standard deviation, 
let $F_{\hat y, \sigma}$ be the CDF of $\hat y$,
then the $\sigma$ that minimizes the CRPS ($\sigma^\star$) is
\begin{gather*}
    \sigma^\star = \argmin_{\sigma > 0} \mathrm{CRPS}(F_{\hat y, \sigma}, y) = \frac{|y|}{\sqrt{\ln 2}}.
\end{gather*}
\end{theorem}
\begin{proof} 
See Appendix \ref{sec:CRPS_opt_std}.
\end{proof}
\begin{corollary}
\label{cor:best_sigma_CRPS}
Given $y$ and $\hat y \sim \mathcal{N}(\mu, \sigma^2)$, the $\sigma$ that minimizes the CRPS ($\sigma^\star$) is
\begin{gather}
\label{eq:coro_3p2}
    \sigma^\star = \argmin_{\sigma > 0} \mathrm{CRPS}(F_{\hat y, \sigma}, y) = \frac{|y-\mu|}{\sqrt{\ln 2}}.
\end{gather}
\end{corollary}
Based on the above corollary, we can calculate the standard deviation that minimizes the CRPS given the ground truth and normally distributed predictions with a fixed mean. However, during inference, the absolute error \(\lvert y - \mu\rvert\) is unknown. In our framework, the final prediction is $z = \hat{y} + \hat{r}$. Since \(\mathscr{M}_{pt,\phi}\) is trained using the MAE loss, we can assume \(\mathbb{E}[\hat{r}]\to 0\). Therefore, we get
\begin{equation}
\label{eq:y_min_mu_is_r}
    z = \hat{y} + \hat{r}\sim \mathcal{N}(\mu,\sigma^2)=\mathcal{N}(\hat{y},\sigma^2)
    \quad\Longrightarrow\quad
    \lvert y - \mu\rvert = \lvert y - \hat{y}\rvert = \lvert r\rvert.
\end{equation}
This shows that one can compute \(\sigma^\star\) from the magnitude of the ground-truth residual \(\lvert r\rvert\). If the residual model effectively captures uncertainty, then \(\lvert r\rvert\) is likely positively correlated with \(\lvert\hat{r}\rvert\), so we approximate
$\lvert r\rvert \approx \alpha\,\mathbb{E}[\lvert\hat{r}\rvert]$,
% \begin{equation}
% \label{eq:approx_r_with_r_hat}
%     \lvert r\rvert \approx \alpha\,\mathbb{E}[\lvert\hat{r}\rvert],
% \end{equation}
where \(\alpha\) is the linear coefficient between \(\lvert r\rvert\) and \(\mathbb{E}[\lvert\hat{r}\rvert]\). 
Combining this approximation with (\ref{eq:coro_3p2}) and (\ref{eq:y_min_mu_is_r}), we obtain
\begin{equation}
\label{eq:eae_final}
    \sigma^\star \approx \frac{\alpha\,\mathbb{E}[\lvert\hat{r}\rvert]}{\sqrt{\ln 2}}.
\end{equation}
(\ref{eq:eae_final}) suggests that the optimal standard deviation is proportional to the expected magnitude of the residuals. This is intuitive: greater residual magnitude implies higher uncertainty, which should be reflected by a larger $\sigma^*$. By choosing an appropriate \(\alpha\) and estimating \(\sigma^\star\) from \(\hat{r}\), we can minimize the CRPS even when the model is trained with the MAE objective. After computing \(\sigma^\star\), we expand or shrink the predictions to match this target variance. We apply this to each $\hat{\mathbf{r}}_{ij} \in \mathbb{R}^{S}$ in $\hat{\mathbf{r}} \in \mathbb{R}^{S \times M \times d}$, where $i \leq M$, $j \leq d$, and each $\hat{\mathbf{r}}_{ij}$ is treated as an empirical Gaussian distribution. Experimental results demonstrate the effectiveness of EAE by comparing it to the use of a fixed expansion factor across all time points (Appendix~\ref{sec:err_exp}). Visualizations are also shown in Appendix~\ref{sec:EAE_viz}.

%%%%%%%%%%%%%%%%%%%%%%%%%%%%%%%%%%%%%%%%%%
\subsubsection{Coverage Optimization (CO)}
\label{sssec:coverage_opt}

Modeling the distribution of the residuals is challenging and trained models might fail to capture the true distribution of the data due to model bias~\cite{NIPS2017_9ef2ed4b}. As a result, a predictor may not have sufficient inductive bias to assign the correct probability to every credible interval. Moreover, due to relatively larger noise in the residuals, overfitting may occur, which is exacerbated by concept drift. In this section, we present the CO algorithm for calibrating the prediction intervals of time series data, achieving an effect similar to that of Kuleshov et al.~\cite{kuleshov2018accurate} but suitable for empirical distributions.

Since the mismatch between the true and predictive distributions can be effectively captured by the Prediction Interval Coverage Probability (PICP) (see detailed definition in Appendix \ref{sec:metrics}), we propose to calibrate the predictive distribution by splitting it into several prediction intervals and finding suitable expansion factors for each interval. Specifically, we define target prediction quantiles to match. 
Let $\boldsymbol{\gamma} = \{\gamma_0, \gamma_1, \cdots, \gamma_l\}$ be a collection of quantiles of the predictive distribution,
where $0 \leq \gamma_i < \gamma_{i+1} < 1, \forall \ 0 \leq i<l$. During the $i$-th step ($i<l$), we adjust the distribution width of all $\hat{r} \in (-\infty, F_{\hat{r}}^{-1}(\frac{1-\gamma_i}{2})] \cup [F_{\hat{r}}^{-1}(\frac{1+\gamma_i}{2}), \infty)$ such that $\mathrm{PICP}(\gamma_{i+1}, r, F_{\hat{r}})=\gamma_{i+1}$. Since $\hat{r} \in \left(F_{\hat{r}}^{-1}\left(\frac{1 - \gamma_i}{2}\right), F_{\hat{r}}^{-1}\left(\frac{1 + \gamma_i}{2}\right)\right)$ is fixed, it follows that the $\mathrm{PICP}(\gamma_j, r, F_{\hat{r}})$ for all $j \leq i$ remain unchanged during the $i$-th step. By incrementing $i$ and adjusting the distribution width, we can calibrate the whole distribution and match the coverage probabilities at any desired prediction interval for the calibration dataset. Detailed visualizations are shown in Appendix~\ref{sec:CO_viz}.

Accordingly, for all $i<l$, we implement binary search and calculate corresponding expansion factors $\boldsymbol{\lambda} = \{\lambda_0, \lambda_1, \cdots, \lambda_l-1\}$ for the adjustment, such that any $\hat{r}$ within range is adjusted as:
\begin{gather}
\label{eq:adjust_rhat_with_lbda}
    \hat{r}_{CO} \leftarrow 
    \begin{cases}
    F_{\hat{r}}^{-1}(\frac{1-\gamma_i}{2}) - \lambda_i(F_{\hat{r}}^{-1}(\frac{1-\gamma_i}{2}) - \hat{r}), & \hat{r} \in (-\infty, F_{\hat{r}}^{-1}(\frac{1-\gamma_i}{2})]; \\
    F_{\hat{r}}^{-1}(\frac{1+\gamma_i}{2}) + \lambda_i(\hat{r} - F_{\hat{r}}^{-1}(\frac{1+\gamma_i}{2})), & \hat{r} \in [F_{\hat{r}}^{-1}(\frac{1+\gamma_i}{2}), \infty).
    \end{cases}
\end{gather}
We use the validation dataset to generate $\boldsymbol{\lambda}$ and apply them during testing in practice. Similar techniques for calibrating using the validation dataset are proposed in several time series analysis  literature~\cite{kuleshov2018accurate,NPSR,lin2022conformal,romano2019conformalized}. Our method stands out as a simple yet effective approach for empirical distributions, without requiring any assumptions about the underlying distribution.

% \ycncmt{Need to point out what is new for your CO method than other similar techniques.}

\subsection{Diffusion Model Network Architecture}
\label{ssec:arch}

In this section, we provide detailed steps as to how the neural network model $\mathscr{M}_{res,\theta}$ in \eqref{eq:cond_noise_pred} denoises $\mathbf{r}^k$ and provides an estimation of the ground truth noise ($\hat{\boldsymbol{\varepsilon}}^{0,k}$). All the concatenation operations are performed sequentially on the first dimension. In particular, given $\mathbf{a} \in \mathbb{R}^{d_a \times D}$ and $\mathbf{b} \in \mathbb{R}^{d_b \times D}$, we have $\verb|concat|(\mathbf{a}, \mathbf{b}): \mathbb{R}^{d_a \times D} \times \mathbb{R}^{d_b \times D} \rightarrow \mathbb{R}^{(d_a + d_b) \times D}$.
Throughout this section, $\mathbf{z}_{i}, \forall i \in \{1,2,\cdots,6\}$ stands for intermediate representations calculated in the network. We concatenate $\mathbf{x}$ and $\hat{\mathbf{y}}$, individually process each variate by standard normalization, then concatenate the result with $\mathbf{r}^k$ such that $\mathbf{z}_1 = \texttt{concat}(\mathbf{x}, \hat{\mathbf{y}}) \in \mathbb{R}^{(N+M) \times d}$ and $\mathbf{z}_2 = \texttt{concat}(\Tilde{\mathbf{z}}_1, \mathbf{r}^k) \in \mathbb{R}^{(N+2M) \times d}$. The resulting embedding $\mathbf{z}_2$ is then transposed, passed through a fully connected layer, and then passed through a $\verb|GELU|$ activation function \cite{GELU}, yielding $\mathbf{z}_3$. We then embed the diffusion step $k$ with two fully connected layers and concatenate it to $\mathbf{z}_3$, yielding $\mathbf{z}_4$:
\begin{gather}
\label{eq:embed_H}
    \mathbf{z}_3 = \texttt{GELU}(\texttt{Linear}(\mathbf{z}_2^\top)) \in \mathbb{R}^{d \times H}, \quad 
    \mathbf{z}_4 = \texttt{concat}(\mathbf{z}_3, \texttt{embed}(k)) \in \mathbb{R}^{(d+d_k) \times H}.
\end{gather}
To this point, all variates are independently processed. To further capture the inter-variate dependencies along with time-dependent relationships, we incorporate bidirectional Mamba layers as
\begin{gather*}
\label{eq:biMamba_layer}
    \mathbf{z}_5 = [\texttt{Mamba}(\mathbf{z}_4^\top) + \texttt{Mamba}(\texttt{Flip}(\mathbf{z}_4^\top))]^\top \in \mathbb{R}^{(d+d_k) \times H}.
\end{gather*}
The usage of a bidirectional Mamba layer stems from the fact that Mamba's selection mechanism can only incorporate antecedent time points \cite{SMamba}. Instead of applying attention in the variate dimension \cite{iTransformer}, we apply attention in the time dimension, which we found to be more effective.

After incorporating the Mamba layer, the output $\mathbf{z}_5$ is projected back to its original length $M$, and the first $d$ variates are extracted, yielding $\mathbf{z}_6 = [\texttt{Linear}(\mathbf{z}_5)^\top]_{0:d} \in \mathbb{R}^{M \times d}.$
Building on top of the assumption that $\mathbf{r}^0$ is normalized yielding a mean of $0$ and variance of $\mathbf{I}$, finally, we add a portion of $\mathbf{r}^k$ to $\mathbf{z}_6$ as the final output: $\hat{\boldsymbol{\varepsilon}}^{0,k} = \mathbf{z}_6 + \mathbf{r}^k k/K \in \mathbb{R}^{M \times d}$.
By calculation, the correlation between $\mathbf{r}^k$ and $\boldsymbol\varepsilon^k$ component-wise is 
\begin{gather}
\label{eq:rho_rk_ek}
    \rho_{r^k, \varepsilon^k} = 
    \frac{\mathrm{cov}(r^k, \varepsilon^k)}{\sqrt{\mathrm{var}(r^k) \mathrm{var}(\varepsilon^k)}} 
    = 
    \mathbb{E}[r^k\varepsilon^k] = 
    \sqrt{1-\bar\alpha_k}\mathbb{E}[(\varepsilon^k)^2] = 
    \sqrt{1-\bar\alpha_k},
\end{gather}
where the third equality holds because $r^0 \indep \varepsilon^k$. From (\ref{eq:forward_sample_k}), we have $\rho_{r^k, \varepsilon^k} \rightarrow 1$ as $k \rightarrow K$. Therefore, adding a scalable portion of $\mathbf{r}^k$ to the final output increases the inductive bias of the model.

\subsection{Training and Inference}
\label{ssec:train_test}

The training algorithm is shown in Algorithm \ref{alg:train}. For each $\mathbf{x}$, we sample a $k$ and $\boldsymbol\varepsilon^k$, and minimize an aggregated loss function for simultaneously optimizing $\mathscr{M}_{pt,\phi}$ and $\mathscr{M}_{res,\theta}$, i.e.,
\begin{gather*}
    \min_{\phi, \theta} \mathscr{L}(\phi, \theta) = \min_{\phi, \theta} \mathbb{E}_{\mathbf{x}, \boldsymbol\varepsilon, k}[\mathscr{L}_{pt}(\phi) + \mathscr{L}_{res}(\theta)],
\end{gather*}
where $\mathscr{L}_{pt}(\phi) = \|\mathbf{y} - \mathscr{M}_{pt,\phi}(\mathbf{x})\|_1$ is the L$_1$-loss between the ground truth future values and the point-based predictions, and $\mathscr{L}_{res}(\theta) = \|\boldsymbol\varepsilon^k - \mathscr{M}_{res,\theta}(\mathbf{r}^k, k, \mathbf{x}, \hat{\mathbf{y}})\|_1$ is the L$_1$-loss between the ground truth residuals and the denoised residuals.
% \begin{gather}
% \label{eq:l_pt}
%     \mathscr{L}_{pt}(\phi) = |\mathbf{y} - \mathscr{M}_{pt,\phi}(\mathbf{x})| \\
% \label{eq:l_res}
%     \mathscr{L}_{res}(\theta) = |\boldsymbol\varepsilon^k - \mathscr{M}_{res,\theta}(\mathbf{r}^k, k, \mathbf{x}, \hat{\mathbf{y}})|
% \end{gather}

The inference algorithm is shown in Algorithm \ref{alg:infer}. For each $\mathbf{x}$, we generate a $\hat{\mathbf{y}}$ using $\mathscr{M}_{pt,\phi}$. During the residual denoising stage, both $\mathbf{x}$ and $\hat{\mathbf{y}}$ are used as the conditioning variable. We generate $\mathbf{r}^K = \boldsymbol\varepsilon^{\kappa_W} \sim \mathcal{N}(0, \mathbf{I})$ and apply (\ref{eq:cond_noise_pred}) and (\ref{eq:r_skip_pred}) repeatedly until $\hat{\mathbf{r}}^0$ is obtained, then de-normalize it using $\boldsymbol\sigma_{trn}$.
After distribution matching (Section \ref{ssec:dist_match}), the final prediction is
\begin{gather}
    \hat{\mathbf{y}}^* = \hat{\mathbf{y}} + \mathbb{E}[\hat{\mathbf{r}}_{CO}] + \boldsymbol\lambda_{EAE} (\hat{\mathbf{r}}_{CO} - \mathbb{E}[\hat{\mathbf{r}}_{CO}]),
\label{eq:final_predictions}
\end{gather}
where the \textit{expand factor} for EAE, $\boldsymbol\lambda_{EAE} = \boldsymbol\sigma^*/\boldsymbol\sigma_{\hat{\mathbf{r}}_{CO}}$, corresponds to the ratio that the distribution needs to be expanded to match the optimal distribution. If no distribution matching is applied, then $\boldsymbol\lambda_{EAE} = 1$ and (\ref{eq:final_predictions}) degrades to $\hat{\mathbf{y}} + \hat{\mathbf{r}_{CO}}$. \\
\begin{minipage}[t]{0.48\textwidth}
\begin{algorithm}[H]
   \caption{Training}
   \label{alg:train}
\begin{algorithmic}
   \REPEAT
    \STATE Sample $\mathbf{x}$ and $\mathbf{y}$ from the training set 
    \STATE $\hat{\mathbf{y}} \leftarrow \mathscr{M}_{pt,\phi}(\mathbf{x})$
    \STATE $\mathbf{r}^0 \leftarrow (\mathbf{y} - \hat{\mathbf{y}})/\boldsymbol\sigma_{trn}$
    \STATE $k \sim \texttt{Uniform}(\{1,2,...,K\})$
    \STATE $\boldsymbol{\varepsilon} \sim \mathcal{N}(0,\mathbf{I})$
    \STATE $\mathbf{r}^k \leftarrow \sqrt{\bar{\alpha}_k} \mathbf{r}^0 + \sqrt{1 - \bar{\alpha}_k} \boldsymbol{\varepsilon}$
    \STATE $\hat{\boldsymbol\varepsilon}^{0,k} \leftarrow \mathscr{M}_{res,\theta}(\mathbf{r}^k, k, \mathbf{x}, \hat{\mathbf{y}})$
    \STATE Calculate $\mathscr{L}_{pt}(\phi)$ and $\mathscr{L}_{res}(\theta)$
    \STATE Do gradient descent on $\nabla_\theta \mathscr{L}_{res}$ and $\nabla_\phi \mathscr{L}_{pt}$
   \UNTIL{converged}
\end{algorithmic}
\end{algorithm}
\end{minipage}%
\hfill
\begin{minipage}[t]{0.48\textwidth}
\begin{algorithm}[H]
   \caption{Inference}
   \label{alg:infer}
\begin{algorithmic}
    \STATE {\bfseries Input:} $\mathbf{x}, \boldsymbol\sigma_{trn}$
    \STATE $\hat{\mathbf{y}} \leftarrow \mathscr{M}_{pt,\phi}(\mathbf{x})$
    \STATE $\mathbf{r}^K \leftarrow \mathcal{N}(0, \mathbf{I})$
    
    \FOR{$i=W$ {\bfseries to} $1$}
    \STATE $\hat{\boldsymbol\varepsilon}^{0,\kappa_i} \leftarrow \mathscr{M}_{res,\theta}(\hat{\mathbf{r}}^{\kappa_i}, \kappa_i, \mathbf{x}, \hat{\mathbf{y}})$

    \STATE Calculate $\hat{\mathbf{r}}^{\kappa_{i-1}}$ using (\ref{eq:r_skip_pred})

    \ENDFOR
    
    \STATE $\hat{\mathbf{r}} \leftarrow \hat{\mathbf{r}}^0\boldsymbol\sigma_{trn}$
    \STATE $\hat{\mathbf{r}}_{CO} \leftarrow$ Adjust $\hat{\mathbf{r}}$ using (\ref{eq:adjust_rhat_with_lbda})
    \STATE $\boldsymbol\lambda_{EAE} \leftarrow \alpha \mathbb{E}[|\hat{\mathbf{r}}_{CO}|] / \boldsymbol\sigma_{\hat{\mathbf{r}}_{CO}}\sqrt{ln 2}$
    
    \STATE $\hat{\mathbf{y}}^* \leftarrow \hat{\mathbf{y}} + \mathbb{E}[\hat{\mathbf{r}}_{CO}] + \boldsymbol\lambda_{EAE}(\hat{\mathbf{r}}_{CO} - \mathbb{E}[\hat{\mathbf{r}}_{CO}]) 
    $
    \STATE \textbf{return} $\hat{\mathbf{y}}^*$
\end{algorithmic}
\end{algorithm}
\end{minipage}

%%%%%%%%%%%%%%%%%%%%%%%%%%%%%%%%%%%%%%%%%%%%%%%%%%%%%%%%%%%%%%%%%%%%%%%%%%%%%%%%%%
%%%%%%%%%%%%%%%%%%%%%%%%%%%%%%%%%%%%%%%%%%%%%%%%%%%%%%%%%%%%%%%%%%%%%%%%%%%%%%%%%%
%%%%%%%%%%%%%%%%%%%%%%%%%%%%%%%%%%%%%%%%%%%%%%%%%%%%%%%%%%%%%%%%%%%%%%%%%%%%%%%%%%
\section{Experiments}
\label{sec:exp}

\subsection{Datasets}
\label{ssec:dset}

\vspace{-2pt}
Experiments are conducted on eight widely used public datasets \cite{SMamba}. The Traffic dataset\footnote{https://pems.dot.ca.gov/} contains hourly road occupancy rates collected from 862 sensors, recorded between January 2015 and December 2016. The Weather dataset\footnote{https://www.bgc-jena.mpg.de/wetter/} contains 21 meteorological indicators collected every 10 minutes during 2020. The Electricity dataset\footnote{https://archive.ics.uci.edu/ml/datasets/ElectricityLoadDiagrams20112014} records the hourly electricity consumption of 321 clients from 2012 to 2014. The Exchange dataset records daily exchange rates of eight countries from 1990 to 2016 \cite{lai2018modeling}. The Solar dataset records the solar power production of 137 PV plants, sampled every 10 minutes during 2006 \cite{lai2018modeling}. The ETT dataset contains data on load and oil temperature collected from electricity transformers between July 2016 and July 2018. We use the subsets ETTh1, recorded hourly, and ETTm1 and ETTm2, recorded every 15 minutes, which capture seven transformer-related factors \cite{zhou2021informer}. The statistics of the datasets are summarized in Table~\ref{tab:datasets}.

\begin{table}[h]
\vspace{-5pt}
\centering
\caption{Datasets used in this work.}
\label{tab:datasets}
\resizebox{0.8\columnwidth}{!}{%
\begin{tabular}{l|llllllll}
\hline
Dataset & Traffic & Weather & Electricity & Exchange & ETTm1 & ETTm2 & ETTh1 & Solar \\ \hline
Variates ($d$) & 862 & 21 & 321 & 8 & 7 & 7 & 7 & 137 \\
Timesteps & 17,544 & 52,696 & 26,304 & 7,588 & 69,680 & 69,680 & 17,420 & 52,560 \\
Frequency & 1 hour & 10 min & 1 hour & 1 day & 15 min & 15 min & 1 hour & 10 min \\ \hline
\end{tabular}%
}
\vspace{-5pt}
\end{table}

% \hfill
% \begin{minipage}[t]{0.48\textwidth}
% \centering
% \captionof{table}{Datasets used in this work.}
% \label{tab:datasets}
% \resizebox{1\columnwidth}{!}{%
% \begin{tabular}{@{}llll@{}}
% \toprule
% Dataset     & Variates ($d$) & Timesteps & Frequency  \\ \midrule
% Traffic     & 862            & 17,544    & 1 hour \\
% Weather     & 21             & 52,696    & 10 min \\
% Electricity & 321            & 26,304    & 1 hour \\
% Exchange    & 8              & 7,588     & 1 day  \\
% ETTm1       & 7              & 69,680    & 15 min \\
% ETTm2       & 7              & 69,680    & 15 min \\
% ETTh1       & 7              & 17,420    & 1 hour \\
% Solar       & 137            & 52,560    & 10 min \\ \bottomrule
% \end{tabular}
% }
% \end{minipage}

\subsection{Baselines}
\label{ssec:base}
% but are three models, so three types of baselines? --> yes, this is a typo
% will move to appendix if main text too long
% \ycncmt{You could consider move this section to the appendix and here simply mention that ``we compare our work with three types of baselines (see details in the appendix).'' }
\vspace{-2pt}
Three types of baselines, comprising a total of ten methods, are compared to RDIT:
\begin{enumerate}[label=(\roman*),itemsep=0mm]
  \item Point-based TSF models \cite{PatchTST,SMamba,iTransformer,TimeDiff,TimeFilter}. These models provide point estimates ($\hat{\mathbf{y}}$) and typically focus on the MSE and MAE. Directly using $\hat{\mathbf{y}}$ and calculating their CRPS will result in a value identical to the MAE. However, this is not a fair comparison, as shown in Appendix \ref{sec:CRPS_opt_std}. We thus \textit{reinforce} the TSF models by adding $\mathcal{N}(0, \boldsymbol{\sigma}_{trn}^2)$ to $\hat{\mathbf{y}}$.

  \item Diffusion-based PTSF models \cite{TMDM,SSSD,tactis2,D3U}. These models focus on generating a distribution that covers as much portion of the ground truth as possible, hence, the CRPS and PICP can be used naturally. 
  
  \item Foundation models \cite{chronos}. These large-scale pretrained models support zero- and few-shot forecasting and deliver probabilistic predictions with calibrated uncertainties.
  
\end{enumerate}

\subsection{Main Results}
\label{ssec:main}

\vspace{-2pt}
We incorporate TimeFilter~\cite{TimeFilter} as the point estimator \(\mathscr{M}_{pt,\phi}\) and the bidirectional Mamba‐based network as the diffusion/denoising model \(\mathscr{M}_{res,\theta}\) in this section. Table~\ref{tab:main_res} shows the CRPS (\ref{eq:CRPS}) and PICP distances (\ref{eq:PICP_dis}) of our method compared to the baselines (full results in Appendix \ref{sec:more_results}). For each metric, RDIT excelled on all but one of the eight datasets, indicating strong adaptability to diverse data distributions and effective capture of both long‐term and short‐term temporal dependencies. One particularly surprising result is the performance of point‐based TSF methods augmented with zero‐mean Gaussians: CRPS achieved the second‐best score on six of eight datasets, and PICP distance achieved the second‐best score on five of eight datasets, often outperforming state‐of‐the‐art PTSF methods. This establishes these enhanced point estimators as strong baselines and demonstrates that Gaussian augmentation is effective for uncertainty modeling. Moreover, these findings highlight that a high‐quality point estimator is critical for producing accurate probabilistic forecasts.

We compare the MAE and MSE of our method and the PTSF models in Table~\ref{tab:MAE_MSE_res} (full results in Appendix~\ref{sec:more_pt_results}). To ensure a fair comparison of the point metrics (cf. Appendix~\ref{sec:MAE_lower_bound}), we first collapse the predictive distributions onto their means and use these as point estimates for evaluation. RDIT still achieves state-of-the-art performance in terms of both MAE and MSE. This demonstrates that RDIT excels not only in PTSF tasks but also in the degenerated point forecasting setting. On the other hand, the results for zero-shot prediction using Chronos suggest that it remains challenging for such models to outperform state-of-the-art supervised learning methods. Visualizations comparing the ground truth and the prediction intervals are shown in Appendix \ref{sec:viz_pred_itvl}.

\begin{table}[ht]
\vspace{-10pt}
\centering
\caption{CRPS and PICP distance of different algorithms for PTSF for eight datasets averaged across prediction lengths 24, 48, and 96. {\bf Bold}: best (lowest) value; {\ul underlined}: second to best.}
\label{tab:main_res}
\resizebox{\columnwidth}{!}{%
\begin{tabular}{@{}P{-1pt}|P{-5pt}|P{-4pt}P{-4pt}|P{-4pt}P{-4pt}|P{-4pt}P{-4pt}|P{-4pt}P{-4pt}|P{-4pt}P{-4pt}|P{-4pt}P{-4pt}|P{-4pt}P{-4pt}|P{-4pt}P{-3pt}@{}}
\toprule
\multicolumn{1}{l|}{} & Dataset & \multicolumn{2}{c|}{Traffic} & \multicolumn{2}{c|}{Weather} & \multicolumn{2}{c|}{Electricity} & \multicolumn{2}{c|}{Exchange} & \multicolumn{2}{c|}{ETTm1} & \multicolumn{2}{c|}{ETTm2} & \multicolumn{2}{c|}{ETTh1} & \multicolumn{2}{c}{Solar} \\ \midrule
Type & Method & CRPS & PICP dis & CRPS & PICP dis & CRPS & PICP dis & CRPS & PICP dis & CRPS & PICP dis & CRPS & PICP dis & CRPS & PICP dis & CRPS & PICP dis \\ \midrule
LLM & {\color[HTML]{38761D} Chronos} & 0.358 & 0.929 & 0.156 & 0.920 & 0.254 & 1.059 & 0.124 & 0.679 & 0.410 & 0.818 & 0.212 & 0.921 & 0.342 & 0.802 & 0.473 & 0.978 \\ \midrule
 & {\color[HTML]{0000FF} TimeDiff} & 0.349 & {\ul 0.114} & 0.545 & 0.465 & 0.276 & {\ul 0.067} & 0.135 & 0.152 & 0.324 & 0.141 & 0.197 & 0.158 & 0.298 & 0.218 & 0.272 & 0.210 \\
 & {\color[HTML]{0000FF} iTransformer} & 0.245 & 0.352 & 0.185 & 0.406 & 0.173 & 0.180 & 0.116 & 0.177 & 0.258 & 0.142 & 0.183 & 0.179 & 0.294 & {\ul 0.178} & 0.207 & 0.304 \\
 & {\color[HTML]{0000FF} SMamba} & 0.233 & 0.351 & 0.184 & 0.410 & 0.172 & 0.337 & 0.126 & 0.171 & 0.258 & {\ul 0.138} & 0.183 & 0.182 & 0.294 & 0.197 & 0.207 & 0.318 \\
 & {\color[HTML]{0000FF} PatchTST} & 0.248 & 0.377 & 0.178 & 0.458 & {\ul 0.162} & 0.200 & 0.120 & {\ul 0.142} & {\ul 0.240} & 0.162 & {\ul 0.175} & 0.185 & 0.299 & 0.189 & 0.179 & 0.342 \\
\multirow{-5}{*}{TSF} & {\color[HTML]{0000FF} TimeFilter} & {\ul 0.207} & 0.370 & 0.171 & 0.414 & {\ul 0.162} & 0.159 & {\ul 0.112} & 0.161 & 0.246 & 0.167 & 0.178 & 0.189 & {\ul 0.289} & 0.202 & 0.186 & 0.332 \\ \midrule
 & {\color[HTML]{FF9900} SSSD} & 0.517 & 1.639 & 0.357 & 1.264 & 0.549 & 1.563 & 1.001 & 1.783 & 0.577 & 1.133 & 1.011 & 1.697 & 0.623 & 0.925 & 0.436 & 1.474 \\
 & {\color[HTML]{FF9900} Tactis-2} & 0.443 & 0.780 & 0.168 & 0.447 & 0.335 & 1.054 & 0.130 & 0.182 & 0.279 & 0.262 & 0.292 & 0.681 & 0.508 & 0.848 & 0.216 & 1.272 \\
 & {\color[HTML]{FF9900} TMDM} & 0.237 & 0.172 & 0.160 & 0.486 & 0.211 & 0.143 & 0.232 & 0.203 & 0.330 & 0.267 & 0.219 & 0.507 & 0.368 & 0.336 & 0.201 & 0.293 \\
 & {\color[HTML]{FF9900} D3U} & 0.213 & \textbf{0.113} & {\ul 0.149} & {\ul 0.180} & 0.184 & 0.108 & 0.129 & 0.371 & 0.248 & 0.167 & 0.193 & {\ul 0.145} & 0.339 & 0.523 & \textbf{0.150} & {\ul 0.129} \\ \cmidrule(l){2-18} 
\multirow{-5}{*}{PTSF} & RDIT & \textbf{0.200} & 0.147 & \textbf{0.132} & \textbf{0.158} & \textbf{0.161} & \textbf{0.065} & \textbf{0.111} & \textbf{0.063} & \textbf{0.239} & \textbf{0.034} & \textbf{0.174} & \textbf{0.095} & \textbf{0.287} & \textbf{0.124} & {\ul 0.178} & \textbf{0.047} \\ \bottomrule
\end{tabular}%
}
\vspace{-5pt}
\end{table}

\begin{table}[ht]
\centering
\caption{MAE and MSE of different algorithms for PTSF for eight datasets averaged across prediction lengths 24, 48, and 96. {\bf Bold}: best (lowest) value; {\ul underlined}: second to best.}
\label{tab:MAE_MSE_res}
\resizebox{1\columnwidth}{!}{%
\begin{tabular}{@{}c|cc|cc|cc|cc|cc|cc|cc|cc@{}}
\toprule
Dataset & \multicolumn{2}{c|}{Traffic} & \multicolumn{2}{c|}{Weather} & \multicolumn{2}{c|}{Electricity} & \multicolumn{2}{c|}{Exchange} & \multicolumn{2}{c|}{ETTm1} & \multicolumn{2}{c|}{ETTm2} & \multicolumn{2}{c|}{ETTh1} & \multicolumn{2}{c}{Solar} \\ \midrule
Method & MAE & MSE & MAE & MSE & MAE & MSE & MAE & MSE & MAE & MSE & MAE & MSE & MAE & MSE & MAE & MSE \\ \midrule
{\color[HTML]{38761D} Chronos} & 0.485 & 0.906 & 0.230 & 0.262 & 0.339 & 0.341 & 0.197 & 0.181 & 0.595 & 0.978 & 0.328 & 0.344 & 0.487 & 0.626 & 0.644 & 1.175 \\
{\color[HTML]{FF9900} SSSD} & 0.571 & 0.879 & 0.444 & 0.387 & 0.643 & 0.662 & 1.118 & 1.928 & 0.709 & 0.919 & 1.141 & 2.325 & 0.774 & 1.114 & 0.574 & 0.585 \\
{\color[HTML]{FF9900} Tactis-2} & 0.564 & 0.846 & 0.220 & 0.167 & 0.411 & 0.365 & 0.174 & 0.068 & 0.372 & 0.372 & 0.360 & 0.293 & 0.625 & 0.973 & 0.270 & 0.220 \\
{\color[HTML]{FF9900} TMDM} & 0.297 & {\ul 0.330} & {\ul 0.195} & 0.170 & 0.275 & 0.185 & 0.318 & 0.181 & 0.420 & 0.455 & 0.272 & 0.208 & 0.473 & 0.529 & 0.251 & 0.204 \\
{\color[HTML]{FF9900} D3U} & {\ul 0.281} & 0.496 & 0.196 & {\ul 0.145} & {\ul 0.245} & {\ul 0.151} & {\ul 0.171} & {\ul 0.062} & {\ul 0.337} & {\ul 0.274} & {\ul 0.254} & {\ul 0.153} & {\ul 0.415} & {\ul 0.389} & {\ul 0.222} & {\ul 0.163} \\
RDIT & \textbf{0.224} & \textbf{0.307} & \textbf{0.159} & \textbf{0.116} & \textbf{0.210} & \textbf{0.112} & \textbf{0.150} & \textbf{0.051} & \textbf{0.325} & \textbf{0.271} & \textbf{0.224} & \textbf{0.129} & \textbf{0.377} & \textbf{0.336} & \textbf{0.203} & \textbf{0.153} \\ \bottomrule
\end{tabular}%
}
\vspace{-10pt}
\end{table}

\subsection{Ablation Study}
\label{ssec:ablation}

Table~\ref{tab:ablation} shows the experimental results by incrementally applying different components of our method and comparing their performance. We start with the point estimates directly generated from $\mathscr{M}_{pt,\phi}$. Since there is no predictive distribution, the PICP distance cannot be measured, and the CRPS is equal to the MAE (Appendix~\ref{sec:CRPS_opt_std}). The 1-step denoise row corresponds to setting $\kappa = \{K\}$, where only one denoising step is taken to predict the residuals. After applying DDIM with ten denoising steps ($\kappa = \{K/10, 2K/10, \cdots, K\}$), we observe improvements in CRPS for most datasets; however, the PICP distance does not consistently improve, indicating that gradual denoising does not ensure the coverage is accurately captured. Adding EAE leads to slight improvements in CRPS, and by further incorporating CO, we achieve the best overall performance across both metrics. These results suggest that each component of our framework contributes to enhancing predictive performance. Additional experiments using SMamba~\cite{SMamba} as the point estimator also showed that applying the RDIT framework results in additive improvement (Appendix \ref{sec:perf_promo}). 

\vspace{-5pt}

\begin{table}[h!]
\centering
\caption{CRPS and PICP distance of ablation study results averaged across prediction lengths 24, 48, 96, 192, 336, and 720. {\bf Bold}: best (lowest) value; {\ul underlined}: second to best.}
\label{tab:ablation}
\resizebox{0.9\columnwidth}{!}{%
\begin{tabular}{@{}P{-2pt}|P{-3pt}P{-3pt}|P{-3pt}P{-3pt}|P{-3pt}P{-3pt}|P{-3pt}P{-3pt}|P{-3pt}P{-3pt}|P{-3pt}P{-3pt}@{}}
\toprule
Dataset & \multicolumn{2}{c|}{Weather} & \multicolumn{2}{c|}{Exchange} & \multicolumn{2}{c|}{ETTm1} & \multicolumn{2}{c|}{ETTm2} & \multicolumn{2}{c|}{ETTh1} & \multicolumn{2}{c}{Solar} \\ \midrule
Features & CRPS & PICP dis & CRPS & PICP dis & CRPS & PICP dis & CRPS & PICP dis & CRPS & PICP dis & CRPS & PICP dis \\ \midrule
Point estimate ($\hat{\mathbf{y}}$) & 0.225 & N/A & 0.311 & N/A & 0.367 & N/A & 0.289 & N/A & 0.412 & N/A & 0.235 & N/A \\
1-step denoise & 0.209 & 0.373 & 0.287 & 0.137 & 0.339 & 0.179 & 0.261 & 0.226 & 0.389 & 0.221 & {\ul 0.207} & {\ul 0.293} \\
DDIM & 0.200 & {\ul 0.233} & 0.225 & 0.141 & 0.278 & 0.194 & 0.226 & 0.201 & 0.314 & 0.225 & 0.209 & 0.368 \\
DDIM+EAE & {\ul 0.196} & 0.303 & {\ul 0.225} & {\ul 0.123} & {\ul 0.277} & {\ul 0.169} & \textbf{0.225} & {\ul 0.179} & \textbf{0.313} & {\ul 0.201} & 0.209 & 0.419 \\
DDIM+EAE+CO & \textbf{0.186} & \textbf{0.148} & \textbf{0.223} & \textbf{0.059} & \textbf{0.275} & \textbf{0.042} & {\ul 0.226} & \textbf{0.087} & {\ul 0.314} & \textbf{0.110} & \textbf{0.198} & \textbf{0.079} \\ \bottomrule
\end{tabular}%
}
\vspace{-10pt}
\end{table}

%%%%%%%%%%%%%%%%%%%%%%%%%%%%%%%%%%%%%%%%%%%%%%%%%%%%%%%%%%%%%%%%%%%%%%%%%%%%%%%%%%
%%%%%%%%%%%%%%%%%%%%%%%%%%%%%%%%%%%%%%%%%%%%%%%%%%%%%%%%%%%%%%%%%%%%%%%%%%%%%%%%%%
%%%%%%%%%%%%%%%%%%%%%%%%%%%%%%%%%%%%%%%%%%%%%%%%%%%%%%%%%%%%%%%%%%%%%%%%%%%%%%%%%%
\section{Conclusion}
\label{sec:conclusion}

\vspace{-5pt}

% No, I think this is fine
 % \ycncmt{but the paper is on the mean due to the normal assumption, do you think we should add a sentence to explain this?}

In conclusion, we present RDIT, a novel framework integrating diffusion processes for probabilistic time series forecasting. It consists of a plug-and-play point-based model that estimates the conditional median, and a residual-based model that estimates the distribution of residuals based on this median, smoothly balancing the workload between the two models. We prove that the CRPS can be further optimized for predictions with a Gaussian distribution and derive the EAE and CO algorithms for distribution matching. An important future research is to extend the current approach to arbitrary predictive distributions beyond the Gaussian. Our results show that RDIT is highly efficient due to the usage of DDIMs and superior across datasets from different fields and a wide range of prediction lengths. By modeling uncertainty effectively, we can potentially improve the modeling of market risk for quantitative trading, accelerate decision-making in critical domains such as healthcare, weather forecasting, and manufacturing, and contribute to enhanced AI reliability.
\vspace{-5pt}

{\small
\bibliographystyle{unsrt}
\bibliography{RDIT_ref}
}

\newpage
\appendix

%%%%%%%%%%%%%%%%%%%%%%%%%%%%%%%%%%%%%%%%%%%%%%%%%%%%%%%%%%%%%%%%%%%%%%%%%%%%%%%%%%
%%%%%%%%%%%%%%%%%%%%%%%%%%%%%%%%%%%%%%%%%%%%%%%%%%%%%%%%%%%%%%%%%%%%%%%%%%%%%%%%%%
%%%%%%%%%%%%%%%%%%%%%%%%%%%%%%%%%%%%%%%%%%%%%%%%%%%%%%%%%%%%%%%%%%%%%%%%%%%%%%%%%%
\newpage
\section{Probabilistic Time Series Forecasting}
\label{sec:PTSF}

Time Series Forecasting (TSF) aims to predict future values of multiple interrelated variables based on their historical observations. Let $\mathbf{x} = \{ \mathbf{x}_1, \mathbf{x}_2, \dots, \mathbf{x}_N \}$ denote the history of the time series, where each $\mathbf{x}_t \in \mathbb{R}^d$ represents the values of $d$ variables at time step $t$, and $N$ is the length of the historical data. The prediction target is a sequence $\mathbf{y} = \{ \mathbf{y}_{1}, \mathbf{y}_{2}, \dots, \mathbf{y}_{M} \}$, where each $\mathbf{y}_t \in \mathbb{R}^d$ corresponds to the future values we aim to forecast over a horizon of length $M$. The TSF problem can be formulated mathematically as finding a function $f$ such that
\begin{gather}
    \mathbf{y} = f(\mathbf{x}),
\end{gather}
where $f: \mathbb{R}^{N \times d} \rightarrow \mathbb{R}^{M \times d}$ maps the historical data $\mathbf{x}$ to the future observations $\mathbf{y}$.

Probabilistic Time Series Forecasting (PTSF) aims to predict the future probability distributions of multiple interrelated variables based on their historical observations, where the prediction targets are defined as $\mathbf{y} = \{ p(\mathbf{y}_{1}), p(\mathbf{y}_{2}), \dots, p(\mathbf{y}_{M}) \}$. The PTSF problem can be formulated mathematically as modeling the conditional distribution of the future observations given the history. Specifically, the goal is to find a function $f$ such that
\begin{gather}
    p(\mathbf{y} | \mathbf{x}) = f(\mathbf{x}),
\end{gather}
where $f: \mathbb{R}^{N \times d} \rightarrow \mathcal{P}(\mathbb{R}^{M \times d})$ maps the historical data $\mathbf{x}$ to the joint probability distribution over the future observations $\mathbf{y}$. Here, $\mathcal{P}(\mathbb{R}^{M \times d})$ denotes the space of probability distributions over $\mathbb{R}^{M \times d}$. This allows for the quantification of uncertainty in the predictions, which is crucial for risk assessment and decision-making in various applications. 

%%%%%%%%%%%%%%%%%%%%%%%%%%%%%%%%%%%%%%%%%%%%%%%%%%%%%%%%%%%%%%%%%%%%%%%%%%%%%%%%%%
%%%%%%%%%%%%%%%%%%%%%%%%%%%%%%%%%%%%%%%%%%%%%%%%%%%%%%%%%%%%%%%%%%%%%%%%%%%%%%%%%%
%%%%%%%%%%%%%%%%%%%%%%%%%%%%%%%%%%%%%%%%%%%%%%%%%%%%%%%%%%%%%%%%%%%%%%%%%%%%%%%%%%
\section{Implementation Details}
\label{sec:implement}

The parameters in our setting are largely inspired by \cite{SMamba,TimeFilter,sun2023difusco}. For all the point estimators (SMamba and TimeFilter), we use the same settings as in the original literature. All experiments were implemented using the PyTorch framework and conducted on an NVIDIA RTX 3090 24GB GPU. The train/validation/test split was maintained at 70/10/20 percent across all experiments. PyTorch Lightning library was employed to streamline the code implementation, with the number of workers set to either 16 or 32. The Adam optimizer with weight decay was chosen for optimization, and early stopping was applied during the training of $\mathscr{M}_{pt,\phi}$ to enhance stability. The detailed parameters for training/testing is shown in Table \ref{tab:implementation}. The superscript of parameter values is the corresponding prediction length.

\begin{table}[h]
\centering
\caption{Implementation settings in this work.}
\label{tab:implementation}
\resizebox{\columnwidth}{!}{%
\begin{tabular}{@{}llcccccccc@{}}
\toprule
\multicolumn{1}{l|}{} & \multicolumn{1}{l|}{} & \multicolumn{8}{c}{Dataset} \\ \midrule
\multicolumn{1}{l|}{Parameter} & \multicolumn{1}{l|}{Naming} & Traffic & Weather & Electricity & Exchange & ETTm1 & ETTm2 & ETTh1 & Solar \\ \midrule
\multicolumn{10}{c}{General} \\ \midrule
\multicolumn{1}{l|}{activation function} & \multicolumn{1}{l|}{} & GELU & GELU & GELU & GELU & GELU & GELU & GELU & GELU \\
\multicolumn{1}{l|}{train batch size} & \multicolumn{1}{l|}{\texttt{batch\_size}} & 16 & 32 & $32^{\leq 48}$, $16^{\geq 96}$ & 32 & 32 & 32 & 32 & 32 \\
\multicolumn{1}{l|}{validation/test batch size} & \multicolumn{1}{l|}{\texttt{test\_batch\_size}} & 1 & 1 & 1 & 1 & 1 & 1 & 1 & 1 \\
\multicolumn{1}{l|}{\# of training epochs} & \multicolumn{1}{l|}{\texttt{num\_epochs}} & 100 & 100 & 100 & 100 & 100 & 100 & 100 & 100 \\ \midrule
\multicolumn{10}{c}{DDIM parameters} \\ \midrule
\multicolumn{1}{l|}{\# of samples for ($S$)} & \multicolumn{1}{l|}{\texttt{samples}} & 50 & 100 & 100 & 100 & 100 & 100 & 100 & 100 \\
\multicolumn{1}{l|}{\# of diffusion steps ($K$)} & \multicolumn{1}{l|}{\texttt{diffusion\_steps}} & 1000 & 1000 & 1000 & 1000 & 1000 & 1000 & 1000 & 1000 \\
\multicolumn{1}{l|}{length of $\kappa$} & \multicolumn{1}{l|}{\texttt{inference\_diffusion\_steps}} & 10 & 10 & 10 & 10 & 10 & 10 & 10 & 10 \\
\multicolumn{1}{l|}{inference schedule} & \multicolumn{1}{l|}{\texttt{inference\_schedule}} & cosine & cosine & cosine & cosine & cosine & cosine & cosine & cosine \\
\multicolumn{1}{l|}{$\beta$ schedule} & \multicolumn{1}{l|}{\texttt{beta\_schedule}} & linear & linear & linear & linear & linear & linear & linear & linear \\ \midrule
\multicolumn{10}{c}{Noise estimation network} \\ \midrule
\multicolumn{1}{l|}{$d_k$ (\ref{eq:embed_H})} & \multicolumn{1}{l|}{\texttt{t\_emb}} & 32 & 32 & 8 & 4 & 4 & 4 & 8 & 32 \\
\multicolumn{1}{l|}{\# of Mamba layers} & \multicolumn{1}{l|}{\texttt{diff\_e\_layers}} & 1 & 1 & 1 & 1 & 1 & 1 & 1 & 1 \\
\multicolumn{1}{l|}{$H$ in $\mathscr{M}_{res,\theta}$ (\ref{eq:embed_H})} & \multicolumn{1}{l|}{\texttt{diff\_d\_model}} & 512 & 512 & $256^{\leq 48}$, $512^{\geq 96}$ & 128 & 128 & 128 & 128 & 512 \\
\multicolumn{1}{l|}{FCN dim in Mamba layers} & \multicolumn{1}{l|}{\texttt{diff\_d\_ff}} & 2048 & 128 & $256^{\leq 48}$, $512^{\geq 96}$ & 128 & 128 & 128 & 128 & 512 \\
\multicolumn{1}{l|}{dropout} & \multicolumn{1}{l|}{\texttt{diff\_dropout}} & 0.3 & 0.5 & $0.5^{\leq 48}$, $0.4^{\geq 96}$ & 0.7 & 0.7 & 0.7 & 0.5 & $0.2^{\leq 96}$, $0.5^{\geq 192}$ \\
\multicolumn{1}{l|}{learning rate} & \multicolumn{1}{l|}{\texttt{diff\_learning\_rate}} & 0.0005 & 0.0005 & 0.0005 & 0.0003 & 0.0003 & 0.0003 & $0.0005^{\leq 336}$, $0.3^{720}$ & $0.0001^{\leq 96}$, $0.0005^{\geq 192}$ \\
\multicolumn{1}{l|}{$\lambda$ weight decay for Adam} & \multicolumn{1}{l|}{\texttt{weight\_decay}} & $10^{-5}$ & $10^{-5}$ & $10^{-5}$ & $10^{-5}$ & $10^{-5}$ & $10^{-5}$ & $10^{-5}$ & $10^{-5}$ \\ \midrule
\multicolumn{10}{c}{Error-aware expansion} \\ \midrule
\multicolumn{1}{l|}{$\alpha$ (\ref{eq:eae_final}} & \multicolumn{1}{l|}{\texttt{alpha}} & \multicolumn{8}{c}{1} \\ \midrule
\multicolumn{10}{c}{Coverage optimization} \\ \midrule
\multicolumn{1}{l|}{$\boldsymbol\gamma$ (Section \ref{sssec:coverage_opt})} & \multicolumn{1}{l|}{} & \multicolumn{8}{c}{$\{0.04, 0.08, \cdots, 1\}$} \\ \bottomrule
\end{tabular}%
}
\end{table}

%%%%%%%%%%%%%%%%%%%%%%%%%%%%%%%%%%%%%%%%%%%%%%%%%%%%%%%%%%%%%%%%%%%%%%%%%%%%%%%%%%
%%%%%%%%%%%%%%%%%%%%%%%%%%%%%%%%%%%%%%%%%%%%%%%%%%%%%%%%%%%%%%%%%%%%%%%%%%%%%%%%%%
%%%%%%%%%%%%%%%%%%%%%%%%%%%%%%%%%%%%%%%%%%%%%%%%%%%%%%%%%%%%%%%%%%%%%%%%%%%%%%%%%%
\newpage
\section{Metrics}
\label{sec:metrics}

The mean absolute error (MAE) is a common metric used to measure the average magnitude of errors between predicted values and actual values in a dataset, without considering their direction (positive or negative). It is the L$_1$ difference between the predicted and true values given by 
\begin{gather}
\label{eqn:MAE}
    \mathrm{MAE}(\mathbf{y}, \hat{\mathbf{y}}) = 
    \|\mathbf{y} - \hat{\mathbf{y}}\|_1,
    %\frac{1}{n} \sum_{i = 1}^{n} |\mathbf{y}_i - \hat{\mathbf{y}}_i|
\end{gather}

% I use $\|\mathbf{y} - \hat{\mathbf{y}}\|_1$
% \ycncmt{do you use $\|\mathbf{y} - \hat{\mathbf{y}}\|_1/n$ or $\|\mathbf{y} - \hat{\mathbf{y}}\|_1$ for MAE?}

where $n$ is the total number of data points, $\mathbf{y} = \{y_1, \dots, y_n\}$ represents the true value and 
$\hat{\mathbf{y}} = \{\hat y_1, \dots, \hat y_n\}$ represents the predicted value.

The mean squared error (MSE) is a commonly used metric in regression tasks. It measures the average squared difference between the predicted values and the true values given by 
\begin{gather}
    \mathrm{MSE}(\mathbf{y}, \hat{\mathbf{y}}) = 
    \|\mathbf{y} - \hat{\mathbf{y}}\|_2^2.
    %\sum_{i = 1}^{n} (\mathbf{y}_i - \hat{\mathbf{y}}_i)^2
\end{gather}
 It is well known that the MSE is sensitive to outliers. 

The Continuous Ranked Probability Score (CRPS) is a statistical measure used to evaluate the accuracy of probabilistic forecasts for continuous variables by comparing the cumulative distribution function (CDF) of the forecast and the actual outcome. The CRPS is defined as
\begin{gather}
\label{eq:CRPS}
    \mathrm{CRPS}(F_{\hat{y}},y) = \int_{-\infty}^{\infty} (F_{\hat{y}}(x) - \mathbbm{1}(x \geq y))^2 dx,
\end{gather}
where $F_{\hat{y}}(\cdot)$ is the forecasted CDF for the variable and $x \in \mathbb{R}$ is the observed value. 
The CRPS is different from MAE and MSE because it considers the whole distribution of the forecasts instead of point estimates.
%CRPS is in a sense just the MSE of the predicted CDF and the true CDF. This metric notably differs from simpler metrics such as MAE because of its asymmetric expression: while the forecasts are probabilistic, the observations are deterministic. Unlike the pinball loss function, the CPRS does not focus on any specific point of the probability distribution but considers the distribution of the forecasts as a whole. \\

The Predictive Interval Coverage Probability (PICP) quantifies the fraction of observed values that fall within a model’s predicted intervals over a dataset. It is computed by averaging binary indicators that signal whether each actual outcome lies inside its associated predictive interval and comparing this empirical rate to the nominal confidence level (e.g., 90 \%).
\begin{gather}
    \mathrm{PICP}(\gamma, \mathbf{y}, \mathbf{F_{\hat{\mathbf{y}}}}) = 
    \frac{1}{n} \sum_{i=1}^{n}
    \mathbbm{1} \left\{ y_i \in \left[F_{\hat{y},i}^{-1} \left(\frac{1-\gamma}{2}\right), F_{\hat{y},i}^{-1}
    \left(\frac{1+\gamma}{2}\right) \right] \right\},
\end{gather}
where $\gamma$ is the desired nominal coverage, $n$ is the total number of data points, $\mathbf{y} = \{y_1, \dots, y_n\}$ represents the true value, $\mathbf{F}_{\hat{\mathbf{y}}} = \{F_{\hat{y},1}, \dots, F_{\hat{y},n}\}$ represents the CDF of the forecasts, and $F_{\hat{y},i}^{-1}$ is the inverse CDF (quantile function) of the predictive distribution for the $i$-th point. A PICP equal to the nominal level indicates perfect calibration, whereas values above or below it denote over-coverage or under-coverage, respectively.

We define the PICP distance metric as the sum of the absolute difference between the PICP and its nominal level for $\gamma \in \{0.5, 0.8, 0.95\}$:
\begin{gather}
\label{eq:PICP_dis}
    \mathrm{PICP \; dis} = \sum_{\gamma \in \{0.5,0.8,0.95\}}
    |\mathrm{PICP}(\gamma, \mathbf{y}, \mathbf{F_{\hat{\mathbf{y}}}}) - \gamma|.
\end{gather}
%\ycncmt{Take the average or no?}
This accounts for multiple prediction intervals simultaneously, making the metric more robust.

%%%%%%%%%%%%%%%%%%%%%%%%%%%%%%%%%%%%%%%%%%%%%%%%%%%%%%%%%%%%%%%%%%%%%%%%%%%%%%%%%%
%%%%%%%%%%%%%%%%%%%%%%%%%%%%%%%%%%%%%%%%%%%%%%%%%%%%%%%%%%%%%%%%%%%%%%%%%%%%%%%%%%
%%%%%%%%%%%%%%%%%%%%%%%%%%%%%%%%%%%%%%%%%%%%%%%%%%%%%%%%%%%%%%%%%%%%%%%%%%%%%%%%%%
\section{Choice of MAE and MSE}
\label{sec:MAE_lower_bound}

When evaluating a predictive distribution using point metrics, it is more appropriate to first take the mean of the samples and then compute the metric, rather than computing the metric directly across all samples of the empirical distribution. Assuming $\mathbf{y} \in \mathbb{R}^{n}$ is the ground truth and $\hat{\mathbf{y}} = \{\hat{\mathbf{y}}_1, \dots, \hat{\mathbf{y}}_S\}$ are $S$ predictions for $\mathbf{y}$, where $\hat{\mathbf{y}} \in \mathbb{R}^{S\times n}$, by the definition of the MAE in \eqref{eqn:MAE}, let $\Bar{\hat{\mathbf{y}}} = \frac{1}{S} \sum_{i=1}^S \hat{\mathbf{y}}_i$, by applying the triangle inequality, we have
%between $y$ and $\hat{\mathbf{y}}$ is defined as
%\begin{gather}
%\label{eqn:MAE}
%    \text{MAE}(y, \hat{\mathbf{y}}) = \frac{1}{n} \sum_{i=1}^n |\hat{y}_i - y|
%\end{gather}
%By applying the triangle inequality, let $\Bar{\hat{\mathbf{y}}} = \frac{1}{n} \sum_{i=1}^n \hat{y}_i$, it is trivial to verify the following:
\begin{equation*}
    \text{MAE}(\hat{\mathbf{y}}, \mathbf{y}) = \frac{1}{S} \sum_{i=1}^S\|\hat{\mathbf{y}}_i - \mathbf{y}\|_1 \geq 
    \|\Bar{\hat{\mathbf{y}}} -  \mathbf{y}\|_1 
    = \text{MAE}\left(\Bar{\hat{\mathbf{y}}}, \mathbf{y}\right).
\end{equation*}
The above relation suggests that the MAE computed between the average of the 
$S$ predictions and the ground truth is tighter than the average MAE computed between each individual prediction and the ground truth. Therefore, we adopt the former approach for evaluating MAE in this paper. Following a similar rationale, we evaluate MSE using the formula $\text{MSE}(\Bar{\hat{\mathbf{y}}}, \mathbf{y}) = \|\Bar{\hat{\mathbf{y}}} -  \mathbf{y}\|^2$.

\newpage
\section{CRPS-optimizing Standard Deviation}
\label{sec:CRPS_opt_std}

Assuming the ground truth is $y$ and that the predictions $\hat y$ follow a zero-mean normal distribution $\hat y \sim \mathcal{N}(0, \sigma^2)$. Then there exists an optimal $\sigma$, denoted by $\sigma^*$, that minimizes the CRPS such that
\begin{equation*}
    \sigma^* = \argmin_{\sigma > 0} \text{CRPS}(F_{\hat{y}, \sigma}, y), 
\end{equation*}
where $F_{\hat{y}, \sigma}$ is the CDF of the forecasted $\hat{y}$, parameterized by $\sigma$, and
\begin{align*}
\text{CRPS}(F_{\hat{y}, \sigma}, y)
   & = \int_{-\infty}^{\infty} \left(\Phi({x}/{\sigma}) - \mathbbm{1}_{x \geq y}\right)^2 dx
    = \int_{-\infty}^{y} \Phi^2({x}/{\sigma}) dx + \int_{y}^{\infty} (1-\Phi({x}/{\sigma}))^2 dx \\
    & = \int_{-\infty}^{y} \Phi^2({x}/{\sigma}) dx + \int_{-\infty}^{-y} \Phi^2({x}/{\sigma}) dx
    = \int_{-\infty}^{y} (1 + \mathbbm{1}_{x \leq -y})\Phi^2({x}/{\sigma}) dx,
\end{align*}
where $\Phi(x)$ is the CDF of the standard normal distribution (recall that
$\Phi(x) = \frac{1}{\sqrt{2\pi}} \int_{-\infty}^{x} e^{-u^2/2} du$).
We obtain $\sigma^\star$ by solving 
\begin{align}
\label{eqn:solve-sigma}
\frac{\partial \text{CRPS}(F_{\hat{y}, \sigma}, y)}{\partial \sigma} = 0,
\end{align}
where 
\begin{align*}
    & \frac{\partial \text{CRPS}(F_{\hat{y}, \sigma}, y)}{\partial \sigma}  \\
    & \quad = - \frac{\sqrt{2}}{\sqrt{\pi}\sigma} 
    \int_{-\infty}^{y} (1 + \mathbbm{1}_{x \leq -y}) \Phi(x/\sigma) \frac{x}{\sigma} e^{-x^2/(2\sigma^2)} dx \\
    & \quad = - \sqrt{\frac{2}{\pi}} \int_{-\infty}^{y/\sigma} 
    (1 + \mathbbm{1}_{\Tilde{x} \leq -y/\sigma}) \Phi(\Tilde{x}) \Tilde{x} e^{-\Tilde{x}^2/2} d\Tilde{x} \\
    & \quad = \sqrt{\frac{2}{\pi}} \left( \Phi(\Tilde{x}) e^{-\Tilde{x}^2/2} \Big|_{-\infty}^{y/\sigma} + \Phi(\Tilde{x}) e^{-\Tilde{x}^2/2} \Big|_{-\infty}^{-y/\sigma} \right) 
     - \frac{1}{\pi} \int_{-\infty}^{y/\sigma} (1 + \mathbbm{1}_{\Tilde{x} \leq -y/\sigma}) e^{-\Tilde{x}^2} d\Tilde{x},
\end{align*}
where the change of variables $\Tilde{x} = x/\sigma$ and integration by parts are applied.
The above derivation combining with \eqref{eqn:solve-sigma} implies 
\begin{align*}
    \sqrt{\frac{2}{\pi}}
    \left( \Phi({y}/{\sigma^\star}) + \Phi(-{y}/{\sigma^\star}) \right) e^{-y^2/(2{\sigma^\star}^2)} 
    & = \frac{1}{\pi}\int_{-\infty}^{y/\sigma^\star} (1 + \mathbbm{1}_{\Tilde{x} \leq -y/\sigma^\star})  e^{-x^2} dx \\
    & = \frac{1}{\sqrt{\pi}} \int_{-\infty}^{y/(\sqrt{2}\sigma^\star)} 
    (1 + \mathbbm{1}_{x' \leq -y/(\sqrt{2}\sigma^\star)}) e^{-x'^2/2} dx' \\
    & = \frac{1}{\sqrt{\pi}} \left( \Phi \left(\frac{y}{\sqrt{2}\sigma^\star}\right) + \Phi\left(-\frac{y}{\sqrt{2}\sigma^\star}\right) \right),
\end{align*}
where the change of variables $x' = \Tilde{x} \sqrt{2}$ is applied.
By using the fact that $\Phi(a) + \Phi(-a) = 1$ for any $a \in \mathbb{R}$, the last display implies 
$\sqrt{2} e^{-y^2/(2{\sigma^\star}^2)} = 1$, which further implies 
$$\sigma^\star = \frac{|y|}{\sqrt{\ln 2}}.$$

The derivation for the case when $\hat y \sim \mathcal{N}(\mu, \sigma^2)$ is similar. For brevity, we omit the details, and the optimal value of $\sigma$, denoted by $\sigma^\star$, is given by
\begin{gather}
\label{eq:opt_sigma_fin}
    \sigma^\star = \frac{|y - \mu|}{\sqrt{\ln 2}} = \frac{|\varepsilon|}{\sqrt{\ln 2}}.
\end{gather}
The above result shows that under the normal assumpion of the predictions, if we know the absolute difference (absolute error $\varepsilon$) between the ground truth ($y$) and the mean of the predictions ($\mu$), we can minimize the CRPS by either \textit{stretching} or \textit{shrinking} the distribution of predictions so that the standard deviation of the distribution matches with the optimal standard deviation.

\begin{figure}[h!]
    \centering
    \includegraphics[width=0.45\linewidth]{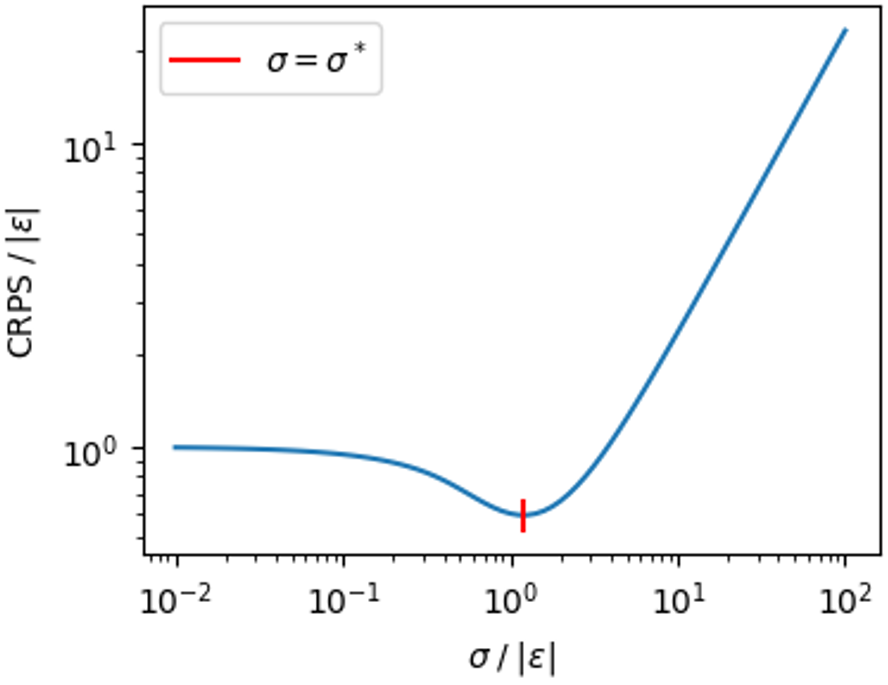}
    \caption{CRPS$/|\varepsilon|$ vs $\sigma / |\varepsilon|$ when the predictions $z$ are drawn from a normal distribution with mean $\mu$ and standard deviation $\sigma$, and the ground truth being $y$.}
    \label{fig:CRPS_vs_STD}
\end{figure}

We verify the theory in Figure \ref{fig:CRPS_vs_STD}. In the figure, we plot the normalized CRPS (CRPS$/|\varepsilon|$) against the normalized standard deviation of the predictions ($\sigma / |\varepsilon|$). According to (\ref{eq:opt_sigma_fin}), $\sigma^\star/|\varepsilon| = 1/\sqrt{\ln 2} \approx 1.2011$, which corresponds to the global minimum indicated by the red line. We also observe in Figure \ref{fig:CRPS_vs_STD} that the CRPS as a function of $\sigma$ is convex: it decreases as $\sigma/|\varepsilon|$ approaches $\sigma^\star$, and then increases rapidly as $\sigma$ departs from $\sigma^\star$. This phenomenon aligns with the experimental results reported in Appendix \ref{sec:err_exp}.
% The phenonnomon suggests that suggesting that using the error-aware expansion is a feasible approach for minimizing the CRPS.

% Reason: if we provide a distribution whose CRPS is larger than the MAE (where the MAE is calculated by first taking the average of the sample means), then we can always further improve the CRPS by shrinking the distribution towards its mean (reduce $\sigma$ to 0), then the CRPS equals the MAE when $\sigma = 0$. If the CRPS of a distribution than the MAE, this suggests that $\sigma$ exceeds a threshold such that the CRPS equals to the MAE.

Let $F_{\hat{y}}(\cdot)$ be the CDF of the predictions for the ground truth $y$. If we condense the predictions to their mean ($\Bar{\hat{y}}$), making the standard deviation of the distribution 0, then $F_{\hat{y}}(x) = \mathbbm{1}\{x \geq \Bar{\hat{y}}\}$, and by the definition of CRPS in \eqref{eq:CRPS},
\begin{gather*}
    \text{CRPS}(F_{\hat{y}}, y) =  \int_{-\infty}^{+\infty} (F_{\hat{y}}(x) - \mathbbm{1}_{x \geq y})^2 \, dx=  \int_{-\infty}^{+\infty} (\mathbbm{1}_{x \geq \Bar{\hat{y}}} - \mathbbm{1}_{x \geq y})^2 \, dx = |\Bar{\hat{y}} - y| = \text{MAE}(\bar{\hat{y}}, y).
\end{gather*}
This relation shows that CRPS equals MAE when the predictive distribution has zero variance (i.e., is concentrated at the mean). For a Gaussian distribution, if the CRPS is larger than the MAE, then based on Figure \ref{fig:CRPS_vs_STD}, the standard deviation of the distribution $\sigma$ must be greater than $\sigma^\star$. This relation can be generalized for a multivariate $\boldsymbol y$ and ${F_{\hat{\mathbf{y}}}}$.

%%%%%%%%%%%%%%%%%%%%%%%%%%%%%%%%%%%%%%%%%%%%%%%%%%%%%%%%%%%%%%%%%%%%%%%%%%%%%%%%%%
%%%%%%%%%%%%%%%%%%%%%%%%%%%%%%%%%%%%%%%%%%%%%%%%%%%%%%%%%%%%%%%%%%%%%%%%%%%%%%%%%%
%%%%%%%%%%%%%%%%%%%%%%%%%%%%%%%%%%%%%%%%%%%%%%%%%%%%%%%%%%%%%%%%%%%%%%%%%%%%%%%%%%
\section{Performance Promotion with Different Point Estimators}
\label{sec:perf_promo}

We use SMamba and TimeFilter as the point estimators, and present results comparing CRPS and MAE before and after applying RDIT (Table \ref{tab:perf_promo}). Across six datasets and averaged over three different prediction lengths, we observe that regardless of the point estimator used, applying RDIT consistently improves the CRPS and often improves the MAE. The reason why point-based metrics such as MAE may not improve with RDIT is straightforward: since RDIT focuses on uncertainty modeling, the mean or median of the predictions is likely to remain unchanged. A similar effect on point-based metrics is also observed in Table~1 of the D3U literature~\cite{D3U}, where equipping a point estimator with uncertainty modeling does not necessarily lead to improved point accuracy.

\begin{table}[h]
\centering
\caption{CRPS and MAE results comparing SMamba and TimeFilter point estimators equipped with a zero-mean Gaussian and RDIT. Results are averaged over prediction lengths 24, 48, and 96.}
\label{tab:perf_promo}
\resizebox{0.9\columnwidth}{!}{%
\begin{tabular}{@{}c|cccc|cccc@{}}
\toprule
Point Model & \multicolumn{4}{c|}{SMamba} & \multicolumn{4}{c}{TimeFilter} \\ \midrule
Metric & \multicolumn{2}{c|}{CRPS} & \multicolumn{2}{c|}{MAE} & \multicolumn{2}{c|}{CRPS} & \multicolumn{2}{c}{MAE} \\ \midrule
Dataset & $\mathcal{N}(0, \boldsymbol\sigma_{trn}^2)$ & \multicolumn{1}{c|}{+RDIT} & $\mathcal{N}(0, \boldsymbol\sigma_{trn}^2)$ & +RDIT & $\mathcal{N}(0, \boldsymbol\sigma_{trn}^2)$ & \multicolumn{1}{c|}{+RDIT} & $\mathcal{N}(0, \boldsymbol\sigma_{trn}^2)$ & +RDIT \\ \midrule
Traffic & 0.233 & \multicolumn{1}{c|}{\textbf{0.204}} & 0.256 & \textbf{0.242} & 0.207 & \multicolumn{1}{c|}{\textbf{0.200}} & 0.228 & \textbf{0.224} \\
Weather & 0.184 & \multicolumn{1}{c|}{\textbf{0.146}} & \textbf{0.174} & 0.179 & 0.171 & \multicolumn{1}{c|}{\textbf{0.132}} & 0.160 & \textbf{0.159} \\
Electricity & 0.172 & \multicolumn{1}{c|}{\textbf{0.169}} & 0.305 & \textbf{0.220} & 0.162 & \multicolumn{1}{c|}{\textbf{0.161}} & \textbf{0.209} & 0.210 \\
Exchange & 0.126 & \multicolumn{1}{c|}{\textbf{0.124}} & \textbf{0.153} & 0.157 & 0.112 & \multicolumn{1}{c|}{\textbf{0.111}} & 0.151 & \textbf{0.150} \\
ETTm2 & 0.183 & \multicolumn{1}{c|}{\textbf{0.182}} & 0.231 & \textbf{0.227} & 0.178 & \multicolumn{1}{c|}{\textbf{0.174}} & 0.225 & \textbf{0.224} \\
Solar & 0.207 & \multicolumn{1}{c|}{\textbf{0.170}} & 0.230 & \textbf{0.210} & 0.186 & \multicolumn{1}{c|}{\textbf{0.178}} & 0.204 & \textbf{0.203} \\ \bottomrule
\end{tabular}%
}
\end{table}

%%%%%%%%%%%%%%%%%%%%%%%%%%%%%%%%%%%%%%%%%%%%%%%%%%%%%%%%%%%%%%%%%%%%%%%%%%%%%%%%%%
%%%%%%%%%%%%%%%%%%%%%%%%%%%%%%%%%%%%%%%%%%%%%%%%%%%%%%%%%%%%%%%%%%%%%%%%%%%%%%%%%%
%%%%%%%%%%%%%%%%%%%%%%%%%%%%%%%%%%%%%%%%%%%%%%%%%%%%%%%%%%%%%%%%%%%%%%%%%%%%%%%%%%
\section{Detailed Results for Probabilistic Metrics}
\label{sec:more_results}

Table~\ref{tab:CRPS_all} shows the CRPS across all prediction lengths, comparing RDIT with the baselines. Table~\ref{tab:PICP_dis_all} presents the corresponding PICP distance results. Together, these tables summarize CRPS and PICP distance performance over a wide range of prediction lengths. From the CRPS results, we observe that RDIT achieves the best performance on average, but the point-based models augmented with zero-mean Gaussians perform comparably—often outperforming recent state-of-the-art PTSF methods, except on the Solar dataset. In terms of PICP distance, these Gaussian-augmented point-based models generally underperform compared to some recent PTSF methods such as D3U. However, TimeDiff stands out among the point-based models, especially on the Electricity dataset. Although TimeDiff is fundamentally a TSF model, its use of diffusion may result in similar training error distributions between training and test datasets, contributing to its strong performance.

\begin{table}[h]
\centering
\caption{CRPS of different algorithms for eight datasets across prediction lengths 24, 48, 96, 192, 336, and 720. {\bf Bold}: best (lowest) value; {\ul underlined}: second to best.}
\label{tab:CRPS_all}
\resizebox{\columnwidth}{!}{%
\begin{tabular}{@{}P{-2.5pt}|P{-2.5pt}|P{-2.5pt}P{-2.5pt}P{-2.5pt}P{-2.5pt}P{-2.5pt}P{-2.5pt}|P{-2.5pt}P{-2.5pt}P{-2.5pt}P{-2.5pt}P{-2.5pt}P{-2.5pt}|P{-2.5pt}P{-2.5pt}P{-2.5pt}P{-2.5pt}P{-2.5pt}P{-2.5pt}|P{-2.5pt}P{-2.5pt}P{-2.5pt}P{-2.5pt}P{-2.5pt}P{-2.5pt}@{}}
\toprule
\multicolumn{1}{l|}{} & Dataset & \multicolumn{6}{c|}{Traffic} & \multicolumn{6}{c|}{Weather} & \multicolumn{6}{c|}{Electricity} & \multicolumn{6}{c}{Exchange} \\ \midrule
Type & Horizon & 24 & 48 & 96 & 192 & 336 & 720 & 24 & 48 & 96 & 192 & 336 & 720 & 24 & 48 & 96 & 192 & 336 & 720 & 24 & 48 & 96 & 192 & 336 & 720 \\ \midrule
LLM & {\color[HTML]{38761D} Chronos} & 0.311 & 0.362 & 0.401 & N/A & N/A & N/A & \textbf{0.104} & 0.155 & 0.208 & N/A & N/A & N/A & 0.226 & 0.254 & 0.283 & N/A & N/A & N/A & 0.082 & 0.115 & 0.174 & N/A & N/A & N/A \\ \midrule
 & {\color[HTML]{0000FF} TimeDiff} & 0.334 & 0.355 & 0.357 & 0.359 & 0.382 & 0.416 & 0.172 & 0.231 & 1.233 & 0.357 & 0.318 & 0.381 & 0.300 & 0.253 & 0.276 & 0.278 & 0.289 & 0.381 & 0.096 & 0.128 & 0.181 & 0.267 & 0.339 & \textbf{0.447} \\
 & {\color[HTML]{0000FF} iTransformer} & 0.265 & 0.234 & 0.237 & 0.245 & 0.250 & 0.269 & 0.156 & 0.190 & 0.210 & 0.245 & 0.280 & 0.320 & 0.160 & 0.176 & 0.185 & 0.201 & 0.214 & 0.228 & 0.081 & 0.111 & 0.155 & 0.224 & {\ul 0.303} & 0.498 \\
 & {\color[HTML]{0000FF} SMamba} & 0.230 & 0.230 & 0.240 & 0.241 & 0.252 & 0.268 & 0.154 & 0.188 & 0.212 & 0.242 & 0.276 & 0.312 & 0.158 & 0.174 & 0.184 & 0.205 & 0.215 & 0.225 & 0.091 & 0.120 & 0.166 & 0.225 & \textbf{0.303} & 0.493 \\
 & {\color[HTML]{0000FF} PatchTST} & 0.237 & 0.250 & 0.258 & 0.264 & 0.266 & 0.283 & 0.146 & 0.177 & 0.212 & 0.239 & 0.269 & 0.371 & \textbf{0.149} & \textbf{0.162} & 0.174 & 0.190 & 0.201 & 0.227 & 0.082 & 0.113 & 0.166 & 0.237 & 0.323 & 0.545 \\
\multirow{-5}{*}{TSF} & {\color[HTML]{0000FF} TimeFilter} & {\ul 0.197} & \textbf{0.207} & {\ul 0.217} & {\ul 0.225} & {\ul 0.233} & {\ul 0.245} & 0.145 & 0.170 & 0.199 & 0.231 & 0.259 & {\ul 0.296} & 0.150 & 0.164 & {\ul 0.173} & {\ul 0.185} & \textbf{0.192} & {\ul 0.208} & {\ul 0.077} & {\ul 0.108} & {\ul 0.152} & {\ul 0.221} & 0.305 & 0.491 \\ \midrule
 & {\color[HTML]{FF9900} SSSD} & 0.459 & 0.535 & 0.556 & 0.660 & 0.709 & 0.789 & 0.313 & 0.386 & 0.373 & 0.446 & 0.433 & 0.450 & 0.487 & 0.582 & 0.576 & 0.635 & 0.614 & 0.753 & 0.626 & 1.150 & 1.228 & 1.141 & 1.195 & 1.184 \\
 & {\color[HTML]{FF9900} Tactis-2} & 0.316 & 0.370 & 0.643 & N/A & N/A & N/A & 0.126 & 0.169 & 0.208 & N/A & N/A & N/A & 0.283 & 0.306 & 0.416 & N/A & N/A & N/A & 0.081 & 0.126 & 0.183 & N/A & N/A & N/A \\
 & {\color[HTML]{FF9900} TMDM} & 0.237 & 0.231 & 0.244 & 0.246 & 0.263 & 0.248 & 0.136 & 0.160 & 0.185 & 0.234 & 0.261 & 0.301 & 0.203 & 0.206 & 0.223 & 0.245 & 0.267 & 0.301 & 0.172 & 0.214 & 0.311 & 0.310 & 0.408 & 0.586 \\
 & {\color[HTML]{FF9900} D3U} & 0.206 & 0.213 & 0.221 & 0.228 & 0.244 & 0.259 & 0.124 & {\ul 0.140} & {\ul 0.182} & {\ul 0.204} & \textbf{0.242} & 0.366 & 0.167 & 0.189 & 0.196 & 0.198 & 0.213 & 0.249 & 0.088 & 0.116 & 0.184 & 0.260 & 0.397 & 0.513 \\
\multirow{-5}{*}{PTSF} & RDIT & \textbf{0.184} & {\ul 0.207} & \textbf{0.209} & \textbf{0.217} & \textbf{0.226} & \textbf{0.241} & {\ul 0.108} & \textbf{0.129} & \textbf{0.158} & \textbf{0.204} & {\ul 0.242} & \textbf{0.273} & {\ul 0.149} & {\ul 0.163} & \textbf{0.170} & \textbf{0.184} & {\ul 0.197} & \textbf{0.207} & \textbf{0.075} & \textbf{0.107} & \textbf{0.151} & \textbf{0.218} & 0.305 & {\ul 0.484} \\ \midrule
\multicolumn{1}{l|}{} & \multicolumn{1}{l|}{} & \multicolumn{6}{c|}{ETTm1} & \multicolumn{6}{c|}{ETTm2} & \multicolumn{6}{c|}{ETTh1} & \multicolumn{6}{c}{Solar} \\ \midrule
LLM & {\color[HTML]{38761D} Chronos} & 0.337 & 0.428 & 0.464 & N/A & N/A & N/A & 0.171 & 0.211 & 0.252 & N/A & N/A & N/A & 0.304 & 0.334 & 0.388 & N/A & N/A & N/A & 0.279 & 0.493 & 0.648 & N/A & N/A & N/A \\ \midrule
 & {\color[HTML]{0000FF} TimeDiff} & 0.286 & 0.341 & 0.344 & 0.365 & 0.399 & 0.413 & 0.165 & 0.198 & 0.229 & 0.267 & 0.302 & 0.344 & 0.279 & 0.297 & 0.319 & 0.345 & 0.370 & 0.375 & 0.211 & 0.274 & 0.332 & 0.371 & 0.406 & 0.408 \\
 & {\color[HTML]{0000FF} iTransformer} & 0.227 & 0.267 & 0.281 & 0.303 & 0.320 & 0.349 & 0.156 & 0.182 & 0.213 & 0.253 & 0.285 & 0.333 & 0.275 & 0.293 & 0.314 & 0.340 & 0.361 & 0.383 & 0.173 & 0.223 & 0.226 & 0.246 & 0.264 & 0.265 \\
 & {\color[HTML]{0000FF} SMamba} & 0.226 & 0.268 & 0.281 & 0.298 & 0.314 & 0.345 & 0.155 & 0.182 & 0.212 & 0.254 & 0.288 & 0.335 & 0.276 & 0.291 & 0.314 & 0.337 & 0.364 & 0.376 & 0.174 & 0.224 & 0.223 & 0.247 & 0.262 & 0.259 \\
 & {\color[HTML]{0000FF} PatchTST} & \textbf{0.210} & {\ul 0.245} & \textbf{0.265} & \textbf{0.287} & \textbf{0.304} & \textbf{0.330} & \textbf{0.148} & {\ul 0.175} & {\ul 0.202} & \textbf{0.235} & {\ul 0.267} & \textbf{0.314} & 0.285 & 0.297 & 0.314 & {\ul 0.329} & {\ul 0.344} & 0.366 & 0.141 & 0.189 & {\ul 0.209} & 0.223 & {\ul 0.230} & 0.244 \\
\multirow{-5}{*}{TSF} & {\color[HTML]{0000FF} TimeFilter} & 0.216 & 0.250 & 0.271 & 0.293 & 0.310 & 0.340 & 0.153 & 0.177 & 0.203 & 0.241 & 0.274 & {\ul 0.323} & {\ul 0.271} & {\ul 0.289} & \textbf{0.307} & \textbf{0.328} & 0.345 & {\ul 0.358} & 0.147 & 0.194 & 0.217 & 0.237 & 0.245 & 0.254 \\ \midrule
 & {\color[HTML]{FF9900} SSSD} & 0.518 & 0.606 & 0.608 & 0.626 & 0.637 & 0.675 & 0.733 & 1.045 & 1.254 & 1.273 & 1.193 & 1.271 & 0.487 & 0.614 & 0.769 & 0.724 & 0.711 & 0.720 & 0.461 & 0.402 & 0.445 & 0.454 & 0.520 & 0.542 \\
 & {\color[HTML]{FF9900} Tactis-2} & 0.251 & 0.295 & 0.291 & N/A & N/A & N/A & 0.170 & 0.298 & 0.409 & N/A & N/A & N/A & 0.305 & 0.549 & 0.669 & N/A & N/A & N/A & 0.184 & 0.211 & 0.253 & N/A & N/A & N/A \\
 & {\color[HTML]{FF9900} TMDM} & 0.280 & 0.343 & 0.366 & 0.383 & 0.422 & 0.445 & 0.177 & 0.230 & 0.251 & 0.335 & 0.474 & 0.565 & 0.351 & 0.364 & 0.389 & 0.421 & 0.500 & 0.516 & 0.166 & 0.220 & 0.217 & 0.213 & 0.250 & 0.231 \\
 & {\color[HTML]{FF9900} D3U} & 0.221 & 0.249 & 0.274 & 0.295 & {\ul 0.305} & {\ul 0.336} & 0.160 & 0.186 & 0.233 & 0.248 & 0.280 & 0.326 & 0.307 & 0.344 & 0.366 & 0.449 & 0.498 & 0.455 & \textbf{0.118} & \textbf{0.170} & \textbf{0.160} & \textbf{0.185} & \textbf{0.189} & \textbf{0.198} \\
\multirow{-5}{*}{PTSF} & RDIT & {\ul 0.210} & \textbf{0.239} & {\ul 0.269} & {\ul 0.289} & 0.306 & 0.339 & {\ul 0.151} & \textbf{0.174} & \textbf{0.197} & {\ul 0.240} & \textbf{0.267} & 0.327 & \textbf{0.271} & \textbf{0.282} & {\ul 0.309} & 0.330 & \textbf{0.341} & \textbf{0.352} & {\ul 0.130} & {\ul 0.187} & 0.217 & {\ul 0.203} & 0.236 & {\ul 0.215} \\ \bottomrule
\end{tabular}%
}
\end{table}

\begin{table}[h]
\centering
\caption{PICP distance of different algorithms for eight datasets across prediction lengths 24, 48, 96, 192, 336, and 720. {\bf Bold}: best (lowest) value; {\ul underlined}: second to best.}
\label{tab:PICP_dis_all}
\resizebox{\columnwidth}{!}{%
\begin{tabular}{@{}P{-2.5pt}|P{-2.5pt}|P{-2.5pt}P{-2.5pt}P{-2.5pt}P{-2.5pt}P{-2.5pt}P{-2.5pt}|P{-2.5pt}P{-2.5pt}P{-2.5pt}P{-2.5pt}P{-2.5pt}P{-2.5pt}|P{-2.5pt}P{-2.5pt}P{-2.5pt}P{-2.5pt}P{-2.5pt}P{-2.5pt}|P{-2.5pt}P{-2.5pt}P{-2.5pt}P{-2.5pt}P{-2.5pt}P{-2.5pt}@{}}
\toprule
\multicolumn{1}{l|}{} & Dataset & \multicolumn{6}{c|}{Traffic} & \multicolumn{6}{c|}{Weather} & \multicolumn{6}{c|}{Electricity} & \multicolumn{6}{c}{Exchange} \\ \midrule
Type & Horizon & 24 & 48 & 96 & 192 & 336 & 720 & 24 & 48 & 96 & 192 & 336 & 720 & 24 & 48 & 96 & 192 & 336 & 720 & 24 & 48 & 96 & 192 & 336 & 720 \\ \midrule
LLM & {\color[HTML]{38761D} Chronos} & 0.828 & 0.853 & 1.106 & N/A & N/A & N/A & 0.771 & 0.832 & 1.157 & N/A & N/A & N/A & 0.982 & 0.990 & 1.205 & N/A & N/A & N/A & 0.516 & 0.558 & 0.964 & N/A & N/A & N/A \\ \midrule
 & {\color[HTML]{0000FF} TimeDiff} & \textbf{0.120} & 0.112 & {\ul 0.110} & {\ul 0.109} & {\ul 0.107} & 0.162 & 0.449 & 0.447 & 0.501 & 0.420 & 0.367 & 0.290 & {\ul 0.085} & {\ul 0.058} & \textbf{0.059} & \textbf{0.054} & {\ul 0.056} & 0.290 & 0.212 & {\ul 0.118} & \textbf{0.125} & {\ul 0.058} & 0.074 & 0.142 \\
 & {\color[HTML]{0000FF} iTransformer} & 0.310 & 0.370 & 0.375 & 0.365 & 0.368 & 0.358 & 0.438 & 0.406 & 0.374 & 0.351 & 0.361 & 0.330 & 0.182 & 0.178 & 0.181 & 0.172 & 0.156 & 0.122 & 0.158 & 0.162 & 0.211 & 0.163 & 0.154 & {\ul 0.096} \\
 & {\color[HTML]{0000FF} SMamba} & 0.293 & 0.380 & 0.379 & 0.392 & 0.361 & 0.377 & 0.441 & 0.424 & 0.366 & 0.351 & 0.367 & 0.344 & 0.668 & 0.163 & 0.180 & 0.166 & 0.132 & 0.113 & 0.147 & 0.162 & 0.204 & 0.168 & 0.151 & \textbf{0.090} \\
 & {\color[HTML]{0000FF} PatchTST} & 0.382 & 0.374 & 0.374 & 0.373 & 0.376 & 0.355 & 0.481 & 0.460 & 0.433 & 0.408 & 0.385 & 0.393 & 0.202 & 0.200 & 0.198 & 0.195 & 0.189 & 0.175 & 0.152 & 0.142 & 0.132 & 0.109 & {\ul 0.064} & 0.271 \\
\multirow{-5}{*}{TSF} & {\color[HTML]{0000FF} TimeFilter} & 0.375 & 0.370 & 0.365 & 0.359 & 0.352 & 0.345 & 0.440 & 0.416 & 0.386 & 0.357 & 0.343 & 0.302 & 0.170 & 0.155 & 0.152 & 0.134 & 0.122 & 0.106 & 0.187 & 0.149 & 0.146 & 0.138 & 0.088 & 0.115 \\ \midrule
 & {\color[HTML]{FF9900} SSSD} & 1.554 & 1.730 & 1.634 & 1.815 & 1.795 & 1.901 & 1.250 & 1.339 & 1.203 & 1.371 & 1.301 & 1.287 & 1.521 & 1.546 & 1.621 & 1.636 & 1.633 & 1.751 & 1.537 & 1.841 & 1.969 & 1.789 & 1.870 & 1.818 \\
 & {\color[HTML]{FF9900} Tactis-2} & 0.692 & 0.450 & 1.197 & N/A & N/A & N/A & 0.320 & 0.840 & 0.183 & N/A & N/A & N/A & 1.205 & 0.185 & 1.771 & N/A & N/A & N/A & 0.179 & 0.125 & 0.243 & N/A & N/A & N/A \\
 & {\color[HTML]{FF9900} TMDM} & {\ul 0.154} & 0.179 & 0.184 & 0.154 & 0.241 & {\ul 0.160} & 0.567 & 0.438 & 0.453 & 0.510 & 0.453 & 0.262 & 0.158 & 0.123 & 0.148 & 0.210 & 0.123 & 0.243 & {\ul 0.083} & 0.128 & 0.399 & 0.415 & 0.609 & 0.466 \\
 & {\color[HTML]{FF9900} D3U} & 0.156 & {\ul 0.104} & \textbf{0.078} & \textbf{0.080} & \textbf{0.093} & \textbf{0.091} & {\ul 0.227} & \textbf{0.227} & \textbf{0.086} & \textbf{0.080} & {\ul 0.112} & {\ul 0.188} & 0.141 & 0.104 & {\ul 0.079} & {\ul 0.071} & 0.077 & \textbf{0.075} & 0.402 & 0.306 & 0.405 & 0.263 & 0.618 & 0.864 \\
\multirow{-5}{*}{PTSF} & RDIT & 0.216 & \textbf{0.088} & 0.136 & 0.151 & 0.123 & 0.264 & \textbf{0.103} & {\ul 0.229} & {\ul 0.141} & {\ul 0.201} & \textbf{0.097} & \textbf{0.118} & \textbf{0.054} & \textbf{0.048} & 0.092 & 0.115 & \textbf{0.056} & {\ul 0.105} & \textbf{0.035} & \textbf{0.026} & {\ul 0.128} & \textbf{0.032} & \textbf{0.015} & 0.116 \\ \midrule
\multicolumn{1}{l|}{} & \multicolumn{1}{l|}{} & \multicolumn{6}{c|}{ETTm1} & \multicolumn{6}{c|}{ETTm2} & \multicolumn{6}{c|}{ETTh1} & \multicolumn{6}{c}{Solar} \\ \midrule
LLM & {\color[HTML]{38761D} Chronos} & 0.693 & 0.744 & 1.017 & N/A & N/A & N/A & 0.845 & 0.829 & 1.088 & N/A & N/A & N/A & 0.710 & 0.706 & 0.989 & N/A & N/A & N/A & 0.839 & 0.883 & 1.212 & N/A & N/A & N/A \\ \midrule
 & {\color[HTML]{0000FF} TimeDiff} & 0.146 & 0.116 & 0.161 & 0.167 & 0.159 & 0.170 & {\ul 0.133} & {\ul 0.149} & 0.193 & {\ul 0.216} & 0.226 & {\ul 0.243} & 0.206 & 0.223 & 0.225 & 0.219 & 0.214 & 0.238 & 0.240 & 0.219 & 0.169 & {\ul 0.213} & {\ul 0.127} & \textbf{0.115} \\
 & {\color[HTML]{0000FF} iTransformer} & 0.139 & 0.134 & 0.151 & 0.192 & 0.204 & 0.205 & 0.150 & 0.168 & 0.220 & 0.250 & 0.265 & 0.272 & {\ul 0.156} & {\ul 0.188} & {\ul 0.192} & 0.190 & 0.198 & {\ul 0.197} & 0.331 & 0.290 & 0.291 & 0.279 & 0.267 & 0.269 \\
 & {\color[HTML]{0000FF} SMamba} & {\ul 0.134} & 0.133 & {\ul 0.147} & 0.209 & 0.210 & 0.195 & 0.142 & 0.167 & 0.238 & 0.265 & 0.270 & 0.266 & 0.187 & 0.202 & 0.203 & 0.194 & {\ul 0.193} & 0.211 & 0.347 & 0.301 & 0.306 & 0.291 & 0.271 & 0.279 \\
 & {\color[HTML]{0000FF} PatchTST} & 0.145 & 0.171 & 0.171 & 0.182 & 0.197 & 0.206 & 0.140 & 0.189 & 0.226 & 0.257 & 0.271 & 0.268 & 0.171 & 0.194 & 0.203 & 0.219 & 0.222 & 0.225 & 0.384 & 0.328 & 0.313 & 0.286 & 0.285 & 0.294 \\
\multirow{-5}{*}{TSF} & {\color[HTML]{0000FF} TimeFilter} & 0.167 & 0.165 & 0.170 & 0.193 & 0.200 & 0.215 & 0.163 & 0.181 & 0.223 & 0.252 & 0.265 & 0.273 & 0.184 & 0.204 & 0.219 & 0.220 & 0.230 & 0.238 & 0.378 & 0.327 & 0.291 & 0.278 & 0.271 & 0.269 \\ \midrule
 & {\color[HTML]{FF9900} SSSD} & 1.097 & 1.216 & 1.087 & 1.078 & 1.171 & 1.184 & 1.508 & 1.781 & 1.802 & 1.708 & 1.537 & 1.655 & 0.787 & 0.918 & 1.068 & 0.885 & 1.090 & 1.260 & 1.673 & 1.491 & 1.260 & 1.551 & 1.478 & 1.279 \\
 & {\color[HTML]{FF9900} Tactis-2} & 0.468 & 0.154 & 0.164 & N/A & N/A & N/A & 0.407 & 0.694 & 0.941 & N/A & N/A & N/A & 0.297 & 0.924 & 1.324 & N/A & N/A & N/A & 0.796 & 1.377 & 1.642 & N/A & N/A & N/A \\
 & {\color[HTML]{FF9900} TMDM} & 0.233 & 0.257 & 0.312 & 0.429 & 0.470 & 0.458 & 0.625 & 0.597 & 0.299 & 0.531 & 0.695 & 0.823 & 0.306 & 0.376 & 0.327 & 0.412 & 0.629 & 0.513 & 0.327 & 0.360 & 0.192 & 0.278 & 0.276 & 0.288 \\
 & {\color[HTML]{FF9900} D3U} & 0.171 & {\ul 0.106} & 0.224 & {\ul 0.083} & {\ul 0.056} & \textbf{0.066} & 0.172 & 0.153 & {\ul 0.110} & 0.261 & {\ul 0.152} & 0.328 & 0.782 & 0.563 & 0.224 & {\ul 0.146} & 0.396 & 0.311 & {\ul 0.064} & {\ul 0.197} & {\ul 0.125} & 0.243 & 0.131 & {\ul 0.130} \\
\multirow{-5}{*}{PTSF} & RDIT & \textbf{0.027} & \textbf{0.031} & \textbf{0.043} & \textbf{0.050} & \textbf{0.036} & {\ul 0.066} & \textbf{0.091} & \textbf{0.092} & \textbf{0.101} & \textbf{0.064} & \textbf{0.047} & \textbf{0.124} & \textbf{0.141} & \textbf{0.169} & \textbf{0.062} & \textbf{0.085} & \textbf{0.094} & \textbf{0.108} & \textbf{0.050} & \textbf{0.018} & \textbf{0.073} & \textbf{0.040} & \textbf{0.066} & 0.228 \\ \bottomrule
\end{tabular}%
}
\end{table}

%%%%%%%%%%%%%%%%%%%%%%%%%%%%%%%%%%%%%%%%%%%%%%%%%%%%%%%%%%%%%%%%%%%%%%%%%%%%%%%%%%
%%%%%%%%%%%%%%%%%%%%%%%%%%%%%%%%%%%%%%%%%%%%%%%%%%%%%%%%%%%%%%%%%%%%%%%%%%%%%%%%%%
%%%%%%%%%%%%%%%%%%%%%%%%%%%%%%%%%%%%%%%%%%%%%%%%%%%%%%%%%%%%%%%%%%%%%%%%%%%%%%%%%%
\newpage
\section{Visualizations of Prediction Intervals}
\label{sec:viz_pred_itvl}

\begin{figure}[h]
    \centering
    \includegraphics[width=0.9\linewidth]{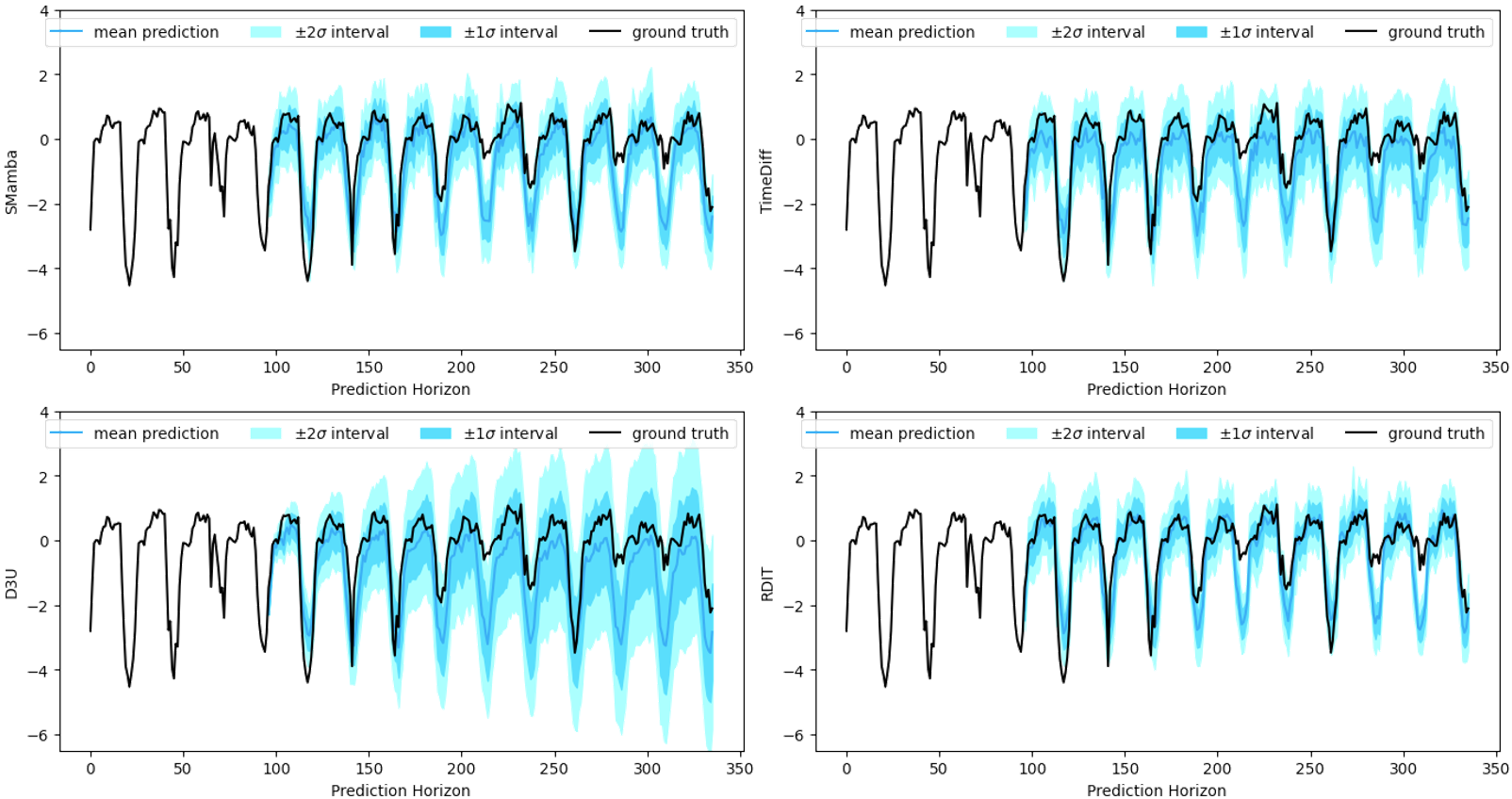}
    \caption{Visualization of prediction intervals for the ETTh1 dataset (variate \#2) using SMamba, TimeDiff, D3U, and RDIT. For each time point \(t\), the mean \(\mu_t\) and standard deviation \(\sigma_t\) are calculated, and two intervals—\([\mu_t - \sigma_t, \mu_t + \sigma_t]\) and \([\mu_t - 2\sigma_t, \mu_t + 2\sigma_t]\)—are plotted over the prediction horizon. The leftmost section of each subfigure, which shows only the ground truth without intervals, represents the input history data provided to the prediction model.}
    \label{fig:ETTh1_pred_itvl}
\end{figure}

\begin{figure}[h]
    \vspace{-10pt}
    \centering
    \includegraphics[width=0.9\linewidth]{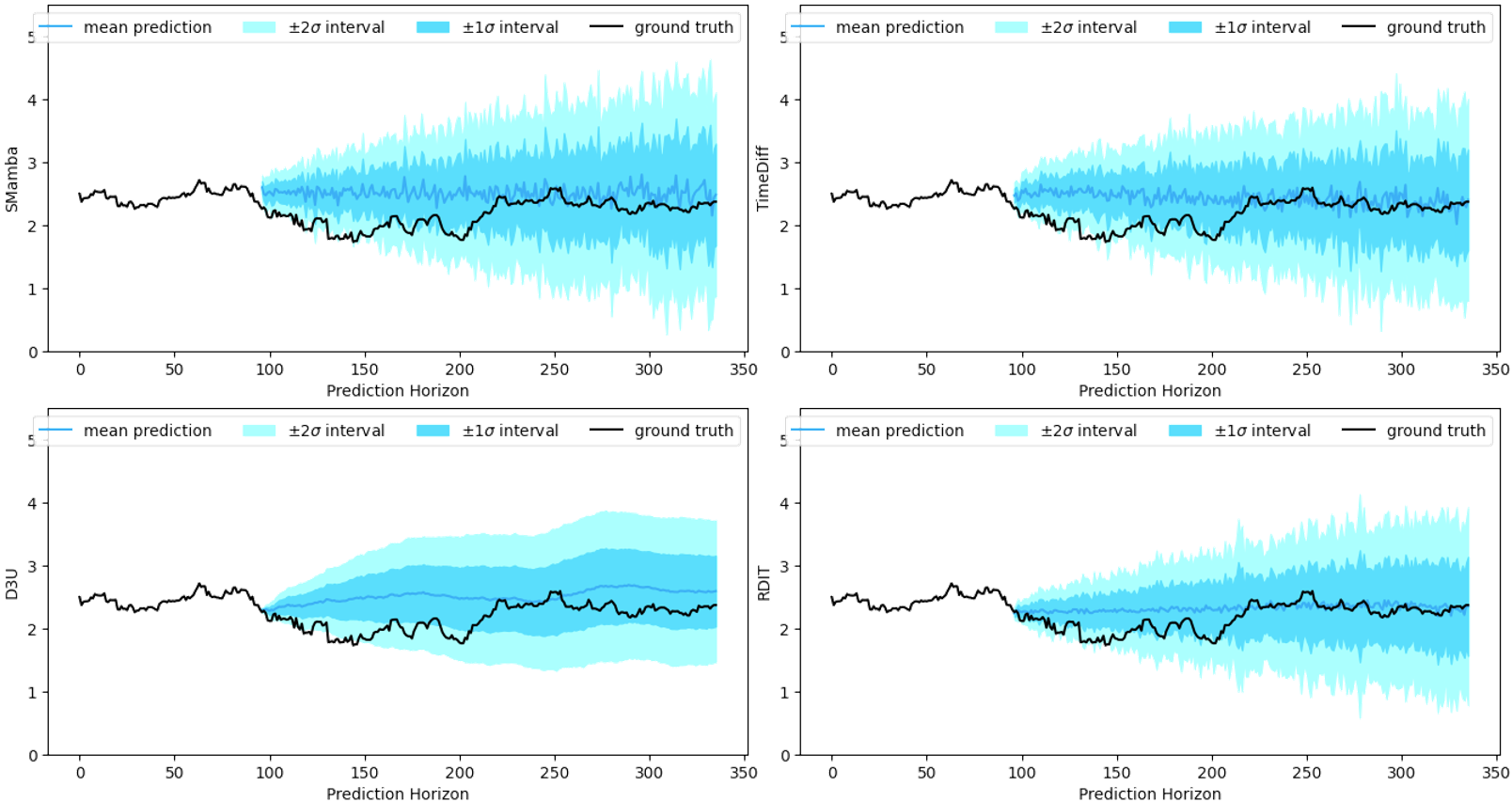}
    \caption{Visualization of prediction intervals for the exchange dataset (variate \#7) using SMamba, TimeDiff, D3U, and RDIT. For each time point \(t\), the mean \(\mu_t\) and standard deviation \(\sigma_t\) are calculated, and two intervals—\([\mu_t - \sigma_t, \mu_t + \sigma_t]\) and \([\mu_t - 2\sigma_t, \mu_t + 2\sigma_t]\)—are plotted over the prediction horizon. The leftmost section of each subfigure, which shows only the ground truth without intervals, represents the input history data provided to the prediction model.}
    \label{fig:exchange_pred_itvl}
    \vspace{-10pt}
\end{figure}

In this section, we visualize the prediction intervals on two datasets to showcase the effectiveness of RDIT for PTSF. Figure~\ref{fig:ETTh1_pred_itvl} visualizes the prediction intervals for the ETTh1 dataset. Although RDIT fails to provide an accurate predictive distribution around time points \(t = \{210, 290, 310\}\), it overall yields better mean estimations and tighter intervals that capture the ground truth, resulting in lower CRPS and PICP distance. Similarly, Figure~\ref{fig:exchange_pred_itvl} shows the prediction intervals for the Exchange dataset. Since the Exchange dataset resembles a random walk, the uncertainty of predictions for all models increases further into the future. Nevertheless, RDIT provides a more stable mean prediction that remains close to the ground truth. Moreover, we observe that SMamba and TimeDiff produce biased estimations starting from the initial prediction point.

%%%%%%%%%%%%%%%%%%%%%%%%%%%%%%%%%%%%%%%%%%%%%%%%%%%%%%%%%%%%%%%%%%%%%%%%%%%%%%%%%%
%%%%%%%%%%%%%%%%%%%%%%%%%%%%%%%%%%%%%%%%%%%%%%%%%%%%%%%%%%%%%%%%%%%%%%%%%%%%%%%%%%
%%%%%%%%%%%%%%%%%%%%%%%%%%%%%%%%%%%%%%%%%%%%%%%%%%%%%%%%%%%%%%%%%%%%%%%%%%%%%%%%%%
% \newpage
\section{Visualizations of Coverage Optimization}
\label{sec:CO_viz}

\begin{figure}[h]
    \centering
    \includegraphics[width=0.9\linewidth]{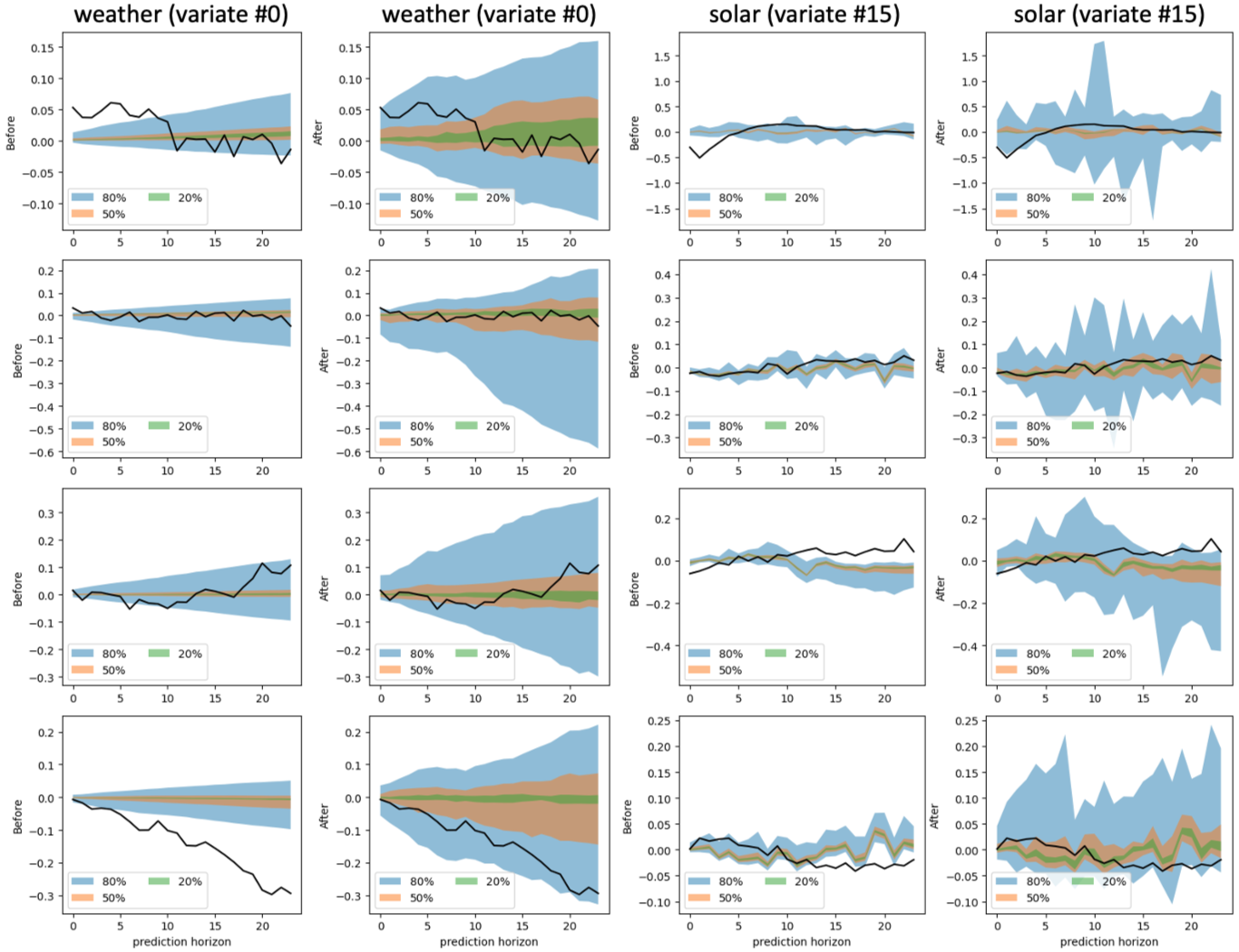}
    \caption{The ground truth and the 20\%, 50\%, and 80\% prediction intervals plotted against the prediction horizon before and after coverage optimization (CO). The first and second columns correspond to the weather dataset before and after CO, respectively, while the third and fourth columns correspond to the solar dataset before and after CO. The plot scales between before and after CO are matched.}
    \label{fig:pred_itvl_CO}
\end{figure}

To provide evidence on why Coverage Optimization (CO) works, we plot the prediction intervals before and after applying CO and compare them to the ground truth, as shown in Figure~\ref{fig:pred_itvl_CO}. Initially, the ground truths are under-covered; that is, the actual observations fall outside the nominal intervals more frequently than expected. After applying CO, the prediction intervals tend to align more closely with the target coverage percentage. For the Solar dataset, CO introduces spikes in the prediction intervals over time, suggesting that while CO improves coverage probability, relying on it alone may negatively impact CRPS. This indicates that further optimization of the prediction distribution may be necessary.

\begin{figure}[h]
    \centering
    \includegraphics[width=0.9\linewidth]{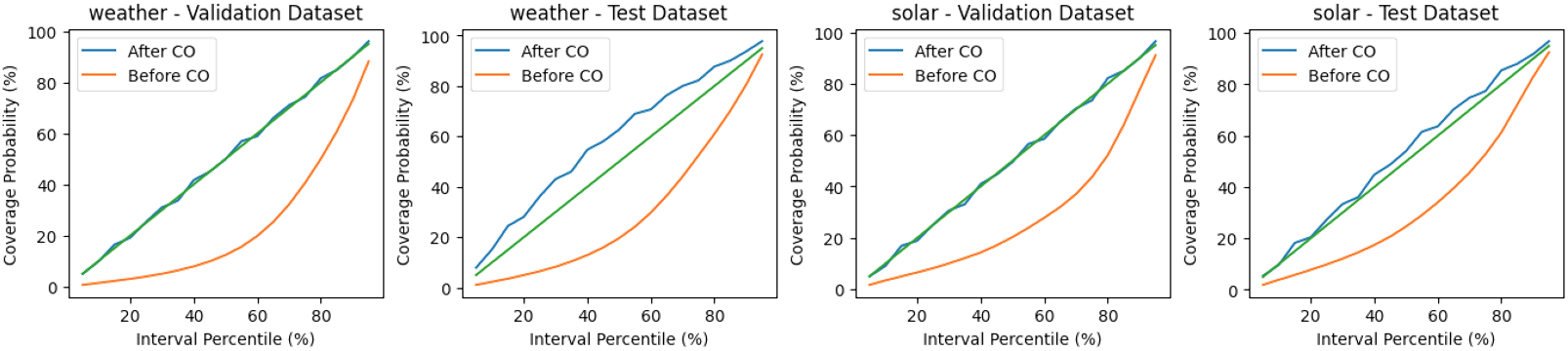}
    \caption{PICP plotted against interval percentile for the weather and solar dataset. The prediction length is 24. The validation datasets are used for calibrating and extracting target expansion factors to be applied to the test dataset.}
    \label{fig:CO_val_test}
\vspace{-5pt}
\end{figure}

Figure~\ref{fig:CO_val_test} shows the PICP at different interval percentiles for the Weather and Solar datasets. By calibrating on the validation dataset and applying the expansion to the test dataset, we can shape the prediction intervals and effectively reduce the mismatch between the true and estimated residual distributions.

%%%%%%%%%%%%%%%%%%%%%%%%%%%%%%%%%%%%%%%%%%%%%%%%%%%%%%%%%%%%%%%%%%%%%%%%%%%%%%%%%%
%%%%%%%%%%%%%%%%%%%%%%%%%%%%%%%%%%%%%%%%%%%%%%%%%%%%%%%%%%%%%%%%%%%%%%%%%%%%%%%%%%
%%%%%%%%%%%%%%%%%%%%%%%%%%%%%%%%%%%%%%%%%%%%%%%%%%%%%%%%%%%%%%%%%%%%%%%%%%%%%%%%%%
\newpage
\section{Error-aware Expansion}
\label{sec:err_exp}

\begin{figure*} [ht]
    \centering
    \includegraphics[width=\linewidth]{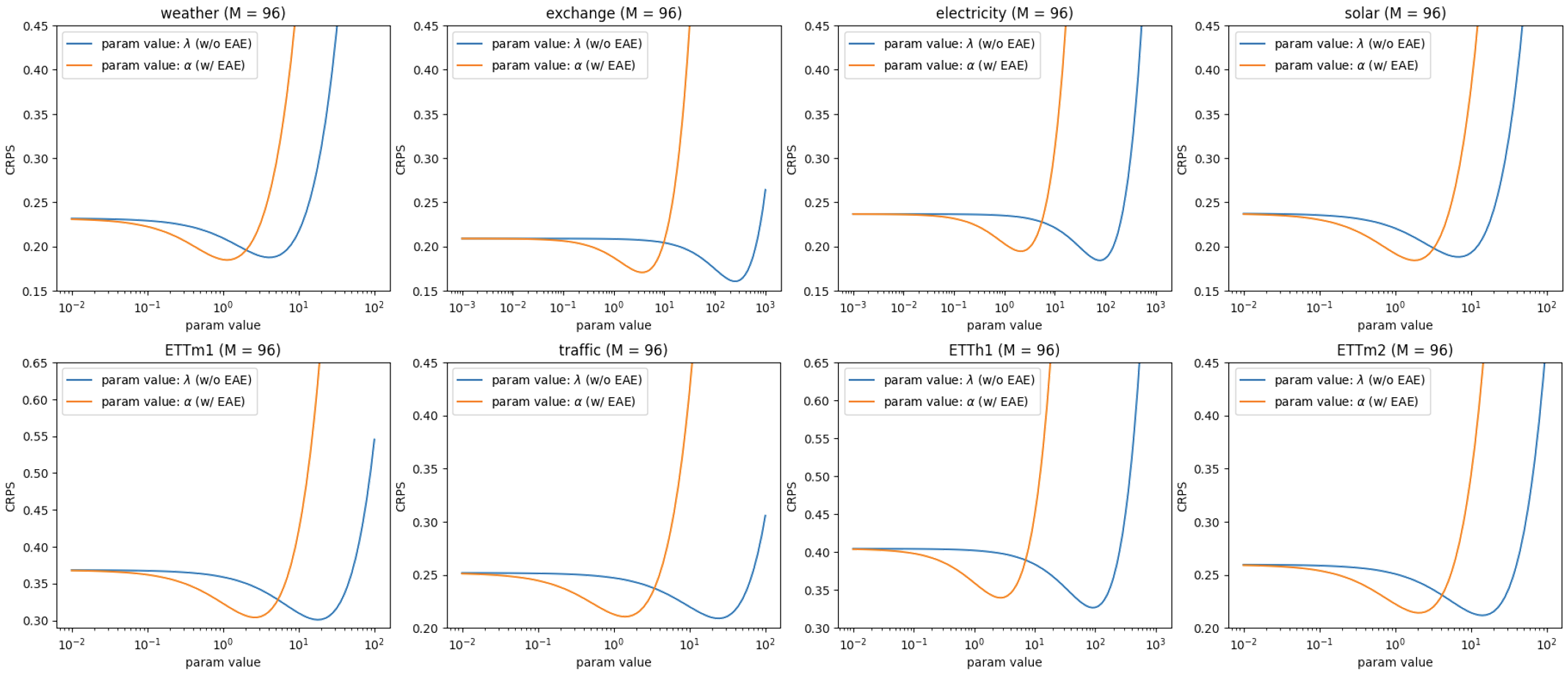}
    \caption{CRPS after applying simple expansion (blue) or EAE (orange) for a prediction length $M = 96$. The x-axis (param value) corresponds to either $\lambda$ (blue) or $\alpha$ (orange). The performance before any expansion is applied—i.e., directly using $\hat{\mathbf{r}}$ as the final residuals—corresponds to $\lambda = 1$ on the blue line.}
    \label{fig:eae_4_odm}
\end{figure*}

The EAE algorithm introduced in Section~\ref{sssec:err_aware_exp} requires access to the residuals in order to estimate the optimal standard deviation. In this section, we aim to answer the following question: Can we simply expand the distribution using a single fixed expansion factor ($\lambda$) to improve performance, potentially even outperforming EAE? Figure~\ref{fig:eae_4_odm} shows the CRPS performance when applying a fixed $\lambda$ for expansion, compared to using EAE. In this experiment, SMamba is used as $\mathscr{M}_{pt,\phi}$. Although simple expansion with $\lambda$ as a hyperparameter may yield better performance in some cases, it requires searching over a wider range of values. Two main advantages of using EAE are evident: (1) the location of the minimum ($\arg\min_{\alpha} \mathrm{CRPS}(\mathbf{F}_{\alpha}, \mathbf{y})$) falls within a relatively narrow range (between 1 and 5), and (2) applying EAE with $\alpha = 1$ consistently improves performance compared to no expansion ($\lambda = 1$). These observations highlight the stability and effectiveness of EAE.

%%%%%%%%%%%%%%%%%%%%%%%%%%%%%%%%%%%%%%%%%%%%%%%%%%%%%%%%%%%%%%%%%%%%%%%%%%%%%%%%%%
%%%%%%%%%%%%%%%%%%%%%%%%%%%%%%%%%%%%%%%%%%%%%%%%%%%%%%%%%%%%%%%%%%%%%%%%%%%%%%%%%%
%%%%%%%%%%%%%%%%%%%%%%%%%%%%%%%%%%%%%%%%%%%%%%%%%%%%%%%%%%%%%%%%%%%%%%%%%%%%%%%%%%
\section{Detailed Prediction Results for Point-based Metrics}
\label{sec:more_pt_results}

Tables~\ref{tab:MAE_all} and \ref{tab:MSE_all} show the MAE and MSE of all probabilistic forecasting models across all prediction lengths. In addition to the probabilistic metrics presented in the previous section, we observe that RDIT also excels in the point-based metrics MAE and MSE when compared with other PTSF methods across multiple prediction horizons. The ranks of the point metrics are more consistent, highlighting the stability of the point estimator, TimeFilter, used in the point prediction stage. Note that due to computational constraints, Chronos and Tactis-2 were not trained or evaluated on prediction lengths 192, 336, and 720.

\begin{table}[h]
\centering
\caption{MAE of different algorithms for eight datasets across prediction lengths 24, 48, 96, 192, 336, and 720. {\bf Bold}: best (lowest) value; {\ul underlined}: second to best.}
\label{tab:MAE_all}
\resizebox{\columnwidth}{!}{%
\begin{tabular}{@{}P{-2.5pt}|P{-2.5pt}|P{-2.5pt}P{-2.5pt}P{-2.5pt}P{-2.5pt}P{-2.5pt}P{-2.5pt}|P{-2.5pt}P{-2.5pt}P{-2.5pt}P{-2.5pt}P{-2.5pt}P{-2.5pt}|P{-2.5pt}P{-2.5pt}P{-2.5pt}P{-2.5pt}P{-2.5pt}P{-2.5pt}|P{-2.5pt}P{-2.5pt}P{-2.5pt}P{-2.5pt}P{-2.5pt}P{-2.5pt}@{}}
\toprule
\multicolumn{1}{l|}{} & Dataset & \multicolumn{6}{c|}{Traffic} & \multicolumn{6}{c|}{Weather} & \multicolumn{6}{c|}{Electricity} & \multicolumn{6}{c}{Exchange} \\ \midrule
Type & Horizon & 24 & 48 & 96 & 192 & 336 & 720 & 24 & 48 & 96 & 192 & 336 & 720 & 24 & 48 & 96 & 192 & 336 & 720 & 24 & 48 & 96 & 192 & 336 & 720 \\ \midrule
LLM & {\color[HTML]{38761D} Chronos} & 0.430 & 0.504 & 0.521 & N/A & N/A & N/A & {\ul 0.161} & 0.242 & 0.287 & N/A & N/A & N/A & 0.308 & 0.347 & 0.363 & N/A & N/A & N/A & 0.143 & 0.198 & 0.251 & N/A & N/A & N/A \\ \midrule
 & {\color[HTML]{FF9900} SSSD} & 0.510 & 0.586 & 0.618 & 0.718 & 0.777 & 0.851 & 0.396 & 0.475 & 0.461 & 0.545 & 0.520 & 0.540 & 0.575 & 0.690 & 0.666 & 0.729 & 0.716 & 0.849 & 0.729 & 1.271 & 1.354 & 1.267 & 1.325 & 1.302 \\
 & {\color[HTML]{FF9900} Tactis-2} & 0.399 & 0.510 & 0.785 & N/A & N/A & N/A & 0.164 & 0.213 & 0.281 & N/A & N/A & N/A & 0.363 & 0.422 & 0.449 & N/A & N/A & N/A & {\ul 0.110} & 0.169 & 0.243 & N/A & N/A & N/A \\
 & {\color[HTML]{FF9900} TMDM} & 0.300 & 0.288 & 0.305 & 0.312 & 0.338 & {\ul 0.312} & 0.162 & 0.196 & {\ul 0.225} & 0.287 & 0.340 & {\ul 0.437} & 0.263 & 0.271 & 0.290 & 0.319 & 0.353 & 0.394 & 0.237 & 0.296 & 0.421 & 0.416 & 0.534 & 0.733 \\
 & {\color[HTML]{FF9900} D3U} & {\ul 0.273} & {\ul 0.281} & {\ul 0.290} & {\ul 0.296} & {\ul 0.324} & 0.333 & 0.161 & {\ul 0.186} & 0.241 & {\ul 0.278} & {\ul 0.320} & 0.487 & {\ul 0.223} & {\ul 0.252} & {\ul 0.261} & {\ul 0.266} & {\ul 0.286} & {\ul 0.333} & 0.117 & {\ul 0.153} & {\ul 0.242} & {\ul 0.344} & {\ul 0.524} & \textbf{0.663} \\
\multirow{-5}{*}{PTSF} & RDIT & \textbf{0.211} & \textbf{0.223} & \textbf{0.239} & \textbf{0.249} & \textbf{0.260} & \textbf{0.275} & \textbf{0.122} & \textbf{0.162} & \textbf{0.194} & \textbf{0.247} & \textbf{0.284} & \textbf{0.339} & \textbf{0.195} & \textbf{0.215} & \textbf{0.221} & \textbf{0.239} & \textbf{0.246} & \textbf{0.274} & \textbf{0.102} & \textbf{0.145} & \textbf{0.205} & \textbf{0.301} & \textbf{0.422} & {\ul 0.690} \\ \midrule
\multicolumn{1}{l|}{} & \multicolumn{1}{l|}{} & \multicolumn{6}{c|}{ETTm1} & \multicolumn{6}{c|}{ETTm2} & \multicolumn{6}{c|}{ETTh1} & \multicolumn{6}{c}{Solar} \\ \midrule
LLM & {\color[HTML]{38761D} Chronos} & 0.500 & 0.638 & 0.647 & N/A & N/A & N/A & 0.273 & 0.343 & 0.369 & N/A & N/A & N/A & 0.442 & 0.492 & 0.525 & N/A & N/A & N/A & 0.410 & 0.696 & 0.827 & N/A & N/A & N/A \\ \midrule
 & {\color[HTML]{FF9900} SSSD} & 0.637 & 0.742 & 0.747 & 0.768 & 0.783 & 0.814 & 0.849 & 1.179 & 1.393 & 1.424 & 1.338 & 1.415 & 0.624 & 0.770 & 0.930 & 0.878 & 0.863 & 0.880 & 0.585 & 0.536 & 0.600 & 0.585 & 0.675 & 0.713 \\
 & {\color[HTML]{FF9900} Tactis-2} & 0.331 & 0.395 & 0.389 & N/A & N/A & N/A & 0.224 & 0.367 & 0.488 & N/A & N/A & N/A & 0.399 & 0.668 & 0.809 & N/A & N/A & N/A & 0.252 & 0.266 & 0.293 & N/A & N/A & N/A \\
 & {\color[HTML]{FF9900} TMDM} & 0.357 & 0.437 & 0.465 & 0.489 & 0.539 & 0.581 & 0.218 & 0.281 & 0.317 & 0.423 & 0.588 & 0.746 & 0.448 & 0.468 & 0.503 & {\ul 0.544} & 0.652 & 0.687 & 0.200 & 0.279 & 0.272 & {\ul 0.268} & 0.317 & {\ul 0.292} \\
 & {\color[HTML]{FF9900} D3U} & {\ul 0.301} & {\ul 0.337} & {\ul 0.373} & {\ul 0.397} & {\ul 0.415} & {\ul 0.455} & {\ul 0.212} & {\ul 0.243} & {\ul 0.308} & {\ul 0.322} & {\ul 0.365} & {\ul 0.417} & {\ul 0.373} & {\ul 0.423} & {\ul 0.450} & 0.564 & {\ul 0.545} & {\ul 0.598} & {\ul 0.176} & {\ul 0.250} & \textbf{0.240} & 0.278 & {\ul 0.284} & 0.298 \\
\multirow{-5}{*}{PTSF} & RDIT & \textbf{0.290} & \textbf{0.330} & \textbf{0.356} & \textbf{0.381} & \textbf{0.404} & \textbf{0.440} & \textbf{0.190} & \textbf{0.226} & \textbf{0.256} & \textbf{0.309} & \textbf{0.344} & \textbf{0.406} & \textbf{0.356} & \textbf{0.367} & \textbf{0.407} & \textbf{0.433} & \textbf{0.447} & \textbf{0.461} & \textbf{0.155} & \textbf{0.199} & {\ul 0.253} & \textbf{0.259} & \textbf{0.277} & \textbf{0.269} \\ \bottomrule
\end{tabular}%
}
\end{table}

\begin{table}[h]
\centering
\caption{MSE of different algorithms for eight datasets across prediction lengths 24, 48, 96, 192, 336, and 720. {\bf Bold}: best (lowest) value; {\ul underlined}: second to best.}
\label{tab:MSE_all}
\resizebox{\columnwidth}{!}{%
\begin{tabular}{@{}P{-2.5pt}|P{-2.5pt}|P{-2.5pt}P{-2.5pt}P{-2.5pt}P{-2.5pt}P{-2.5pt}P{-2.5pt}|P{-2.5pt}P{-2.5pt}P{-2.5pt}P{-2.5pt}P{-2.5pt}P{-2.5pt}|P{-2.5pt}P{-2.5pt}P{-2.5pt}P{-2.5pt}P{-2.5pt}P{-2.5pt}|P{-2.5pt}P{-2.5pt}P{-2.5pt}P{-2.5pt}P{-2.5pt}P{-2.5pt}@{}}
\toprule
\multicolumn{1}{l|}{} & Dataset & \multicolumn{6}{c|}{Traffic} & \multicolumn{6}{c|}{Weather} & \multicolumn{6}{c|}{Electricity} & \multicolumn{6}{c}{Exchange} \\ \midrule
Type & Horizon & 24 & 48 & 96 & 192 & 336 & 720 & 24 & 48 & 96 & 192 & 336 & 720 & 24 & 48 & 96 & 192 & 336 & 720 & 24 & 48 & 96 & 192 & 336 & 720 \\ \midrule
LLM & {\color[HTML]{38761D} Chronos} & 0.777 & 0.963 & 0.979 & N/A & N/A & N/A & 0.178 & 0.289 & 0.317 & N/A & N/A & N/A & 0.287 & 0.359 & 0.377 & N/A & N/A & N/A & 0.124 & 0.208 & 0.211 & N/A & N/A & N/A \\ \midrule
 & {\color[HTML]{FF9900} SSSD} & 0.768 & 0.859 & 1.010 & 1.333 & 1.347 & 1.742 & 0.310 & 0.433 & 0.419 & 0.542 & 0.537 & 0.573 & 0.554 & 0.731 & 0.702 & 0.887 & 0.769 & 1.069 & 0.842 & 2.382 & 2.560 & 2.412 & 2.672 & 2.538 \\
 & {\color[HTML]{FF9900} Tactis-2} & 0.562 & 0.699 & 1.277 & N/A & N/A & N/A & 0.144 & 0.164 & {\ul 0.191} & N/A & N/A & N/A & 0.377 & 0.344 & 0.374 & N/A & N/A & N/A & {\ul 0.026} & 0.060 & 0.118 & N/A & N/A & N/A \\
 & {\color[HTML]{FF9900} TMDM} & {\ul 0.324} & {\ul 0.306} & {\ul 0.361} & {\ul 0.371} & {\ul 0.383} & \textbf{0.367} & 0.144 & 0.174 & 0.192 & 0.247 & 0.368 & 0.515 & 0.177 & 0.170 & 0.209 & 0.252 & 0.282 & 0.345 & 0.094 & 0.147 & 0.301 & 0.311 & 0.517 & 0.976 \\
 & {\color[HTML]{FF9900} D3U} & 0.454 & 0.510 & 0.523 & 0.527 & 0.549 & 0.784 & {\ul 0.107} & {\ul 0.131} & 0.197 & {\ul 0.224} & {\ul 0.294} & {\ul 0.478} & {\ul 0.126} & {\ul 0.160} & {\ul 0.167} & {\ul 0.175} & {\ul 0.198} & {\ul 0.258} & 0.029 & {\ul 0.049} & {\ul 0.109} & {\ul 0.225} & {\ul 0.484} & \textbf{0.767} \\
\multirow{-5}{*}{PTSF} & RDIT & \textbf{0.285} & \textbf{0.305} & \textbf{0.333} & \textbf{0.350} & \textbf{0.368} & {\ul 0.398} & \textbf{0.088} & \textbf{0.114} & \textbf{0.146} & \textbf{0.192} & \textbf{0.238} & \textbf{0.314} & \textbf{0.096} & \textbf{0.115} & \textbf{0.126} & \textbf{0.144} & \textbf{0.143} & \textbf{0.178} & \textbf{0.022} & \textbf{0.044} & \textbf{0.086} & \textbf{0.181} & \textbf{0.345} & {\ul 0.849} \\ \midrule
\multicolumn{1}{l|}{} & \multicolumn{1}{l|}{} & \multicolumn{6}{c|}{ETTm1} & \multicolumn{6}{c|}{ETTm2} & \multicolumn{6}{c|}{ETTh1} & \multicolumn{6}{c}{Solar} \\ \midrule
LLM & {\color[HTML]{38761D} Chronos} & 0.726 & 1.106 & 1.103 & N/A & N/A & N/A & 0.235 & 0.381 & 0.416 & N/A & N/A & N/A & 0.539 & 0.638 & 0.700 & N/A & N/A & N/A & 0.582 & 1.336 & 1.608 & N/A & N/A & N/A \\ \midrule
 & {\color[HTML]{FF9900} SSSD} & 0.808 & 0.952 & 0.998 & 1.051 & 1.071 & 1.174 & 1.291 & 2.376 & 3.309 & 3.417 & 3.197 & 3.485 & 0.749 & 1.074 & 1.521 & 1.482 & 1.293 & 1.253 & 0.642 & 0.520 & 0.592 & 0.597 & 0.707 & 0.759 \\
 & {\color[HTML]{FF9900} Tactis-2} & 0.286 & 0.432 & 0.400 & N/A & N/A & N/A & 0.128 & 0.285 & 0.464 & N/A & N/A & N/A & 0.382 & 1.127 & 1.410 & N/A & N/A & N/A & 0.204 & 0.217 & 0.239 & N/A & N/A & N/A \\
 & {\color[HTML]{FF9900} TMDM} & 0.312 & 0.475 & 0.577 & 0.553 & 0.678 & 0.689 & 0.117 & 0.266 & 0.242 & 0.449 & 0.944 & 0.947 & 0.488 & 0.508 & 0.592 & 0.657 & 0.797 & 0.860 & 0.140 & 0.221 & 0.250 & 0.256 & 0.296 & 0.303 \\
 & {\color[HTML]{FF9900} D3U} & \textbf{0.221} & \textbf{0.276} & {\ul 0.325} & {\ul 0.382} & {\ul 0.399} & {\ul 0.483} & {\ul 0.105} & {\ul 0.139} & {\ul 0.213} & {\ul 0.252} & {\ul 0.314} & {\ul 0.415} & {\ul 0.329} & {\ul 0.394} & {\ul 0.445} & {\ul 0.620} & {\ul 0.612} & {\ul 0.658} & {\ul 0.113} & {\ul 0.187} & \textbf{0.190} & {\ul 0.229} & {\ul 0.244} & {\ul 0.269} \\
\multirow{-5}{*}{PTSF} & RDIT & {\ul 0.223} & {\ul 0.278} & \textbf{0.312} & \textbf{0.356} & \textbf{0.390} & \textbf{0.452} & \textbf{0.092} & \textbf{0.130} & \textbf{0.166} & \textbf{0.250} & \textbf{0.301} & \textbf{0.408} & \textbf{0.301} & \textbf{0.318} & \textbf{0.388} & \textbf{0.429} & \textbf{0.456} & \textbf{0.452} & \textbf{0.097} & \textbf{0.147} & {\ul 0.215} & \textbf{0.222} & \textbf{0.233} & \textbf{0.247} \\ \bottomrule
\end{tabular}%
}
\end{table}

%%%%%%%%%%%%%%%%%%%%%%%%%%%%%%%%%%%%%%%%%%%%%%%%%%%%%%%%%%%%%%%%%%%%%%%%%%%%%%%%%%
%%%%%%%%%%%%%%%%%%%%%%%%%%%%%%%%%%%%%%%%%%%%%%%%%%%%%%%%%%%%%%%%%%%%%%%%%%%%%%%%%%
%%%%%%%%%%%%%%%%%%%%%%%%%%%%%%%%%%%%%%%%%%%%%%%%%%%%%%%%%%%%%%%%%%%%%%%%%%%%%%%%%%
\section{Change of CRPS in Denoising}
\label{sec:CRPS_vs_denoise_step}

\begin{figure}[h]
    \centering
    \includegraphics[width=\linewidth]{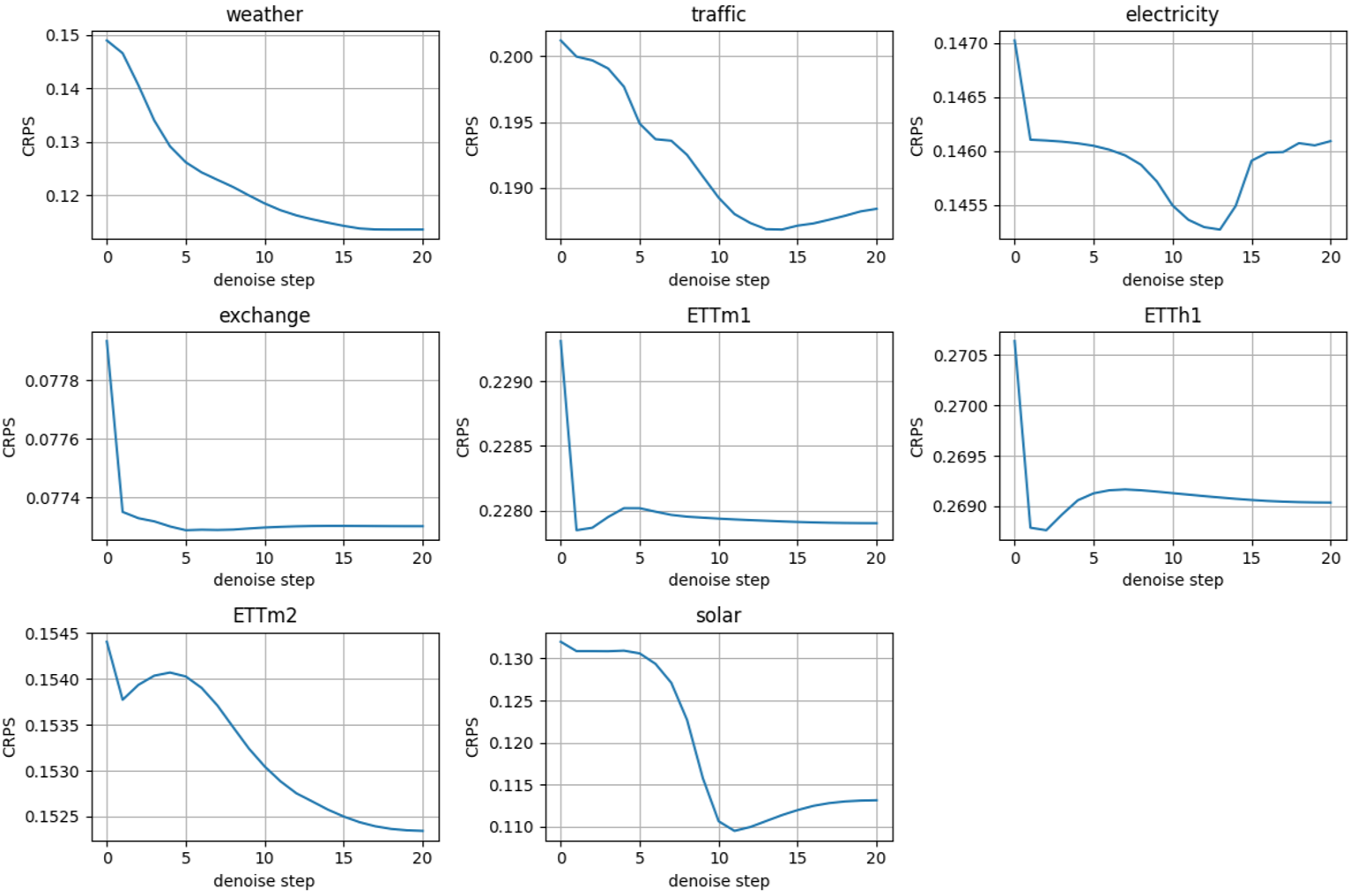}
    \caption{CRPS during each denoising step across eight datasets. The prediction length is 24. The starting point corresponds to a point estimator equipped with $\mathcal{N}(0, \boldsymbol\sigma_{trn}^2)$. EAE and CO were not applied in these experiments.}
    \label{fig:CRPS_denoise}
\end{figure}

Recall that $\hat{\mathbf{r}}^{\kappa_W} \sim \mathcal{N}(0, \mathbf{I})$ (Section \ref{sec:prelim}). This means that if we directly take $\hat{\mathbf{r}}^{\kappa_W}$ and de-normalize it with $\boldsymbol{\sigma}_{trn}$ to match the standard deviation of the real residuals, then add it back to $\hat{\mathbf{y}}$, the result will be equivalent to equipping a point estimator with a zero-mean Gaussian—the strong baseline for reinforced TSF models. We show in Figure~\ref{fig:CRPS_denoise} that during the denoising process, the CRPS can improve \textit{over} this baseline and yield a better final CRPS, with consistent improvements observed across all datasets. Despite these gains, for the Electricity, Exchange, ETTm1, and ETTh1 datasets, the CRPS improved by less than 1\%, indicating that $\boldsymbol{\sigma}_{trn}$ already captures the uncertainty effectively compared to state-of-the-art PTSF methods. Moreover, in the Traffic, Electricity, Exchange, ETTm1, ETTh1, and Solar datasets, we observe that the CRPS of the final denoised step is worse than that of some intermediate steps. This suggests that the distribution of the residuals tends to overfit the training data, implying that the raw denoised results are suboptimal and further optimization, such as distribution matching, is necessary.

%%%%%%%%%%%%%%%%%%%%%%%%%%%%%%%%%%%%%%%%%%%%%%%%%%%%%%%%%%%%%%%%%%%%%%%%%%%%%%%%%%
%%%%%%%%%%%%%%%%%%%%%%%%%%%%%%%%%%%%%%%%%%%%%%%%%%%%%%%%%%%%%%%%%%%%%%%%%%%%%%%%%%
%%%%%%%%%%%%%%%%%%%%%%%%%%%%%%%%%%%%%%%%%%%%%%%%%%%%%%%%%%%%%%%%%%%%%%%%%%%%%%%%%%
\section{Visualization of the Effect of Error-aware Expansion}
\label{sec:EAE_viz}

\begin{figure}[ht]
    \centering
    \includegraphics[width=1\linewidth]{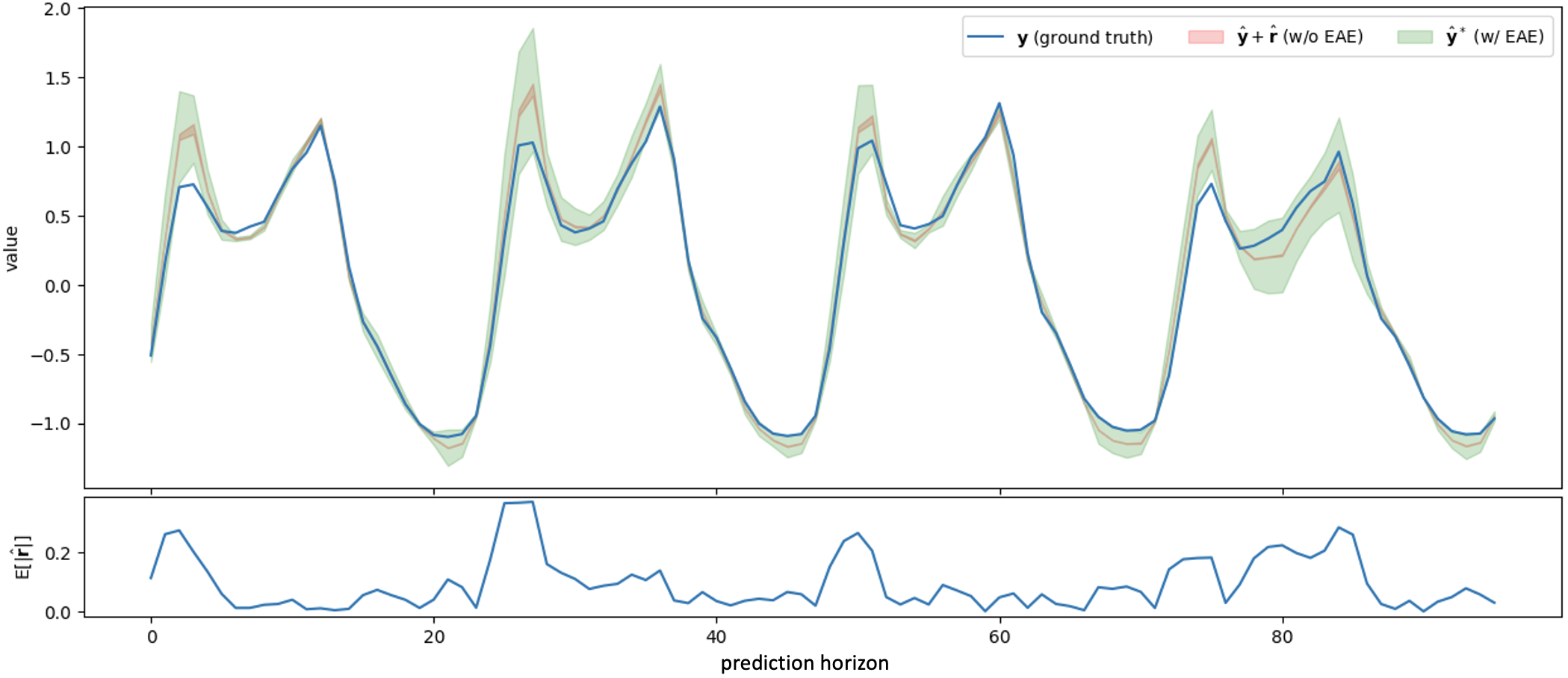}
    \caption{Visualization on the Traffic dataset (variate \#800) with a prediction length of 96. SMamba is used as \(\mathscr{M}_{pt,\phi}\), and the prediction mean is not confined to 0 in this experiment. Coverage Optimization (CO) is not applied. The shaded prediction area corresponds to the \(\pm 1\boldsymbol\sigma\) interval. The top plot shows the ground truth and predictions before and after applying EAE, while the bottom plot displays the corresponding mean of the absolute residuals (\(\mathbb{E}[|\hat{\mathbf{r}}|]\)).}
    \label{fig:EAE_viz}
\end{figure}

Error-aware Expansion (EAE) appropriately assigns a wider distribution to prediction regions with higher uncertainty. Indeed, an important aspect of PTSF is that the predictions should be uncertainty-aware. We provide visualizations as to how this works. Figure \ref{fig:EAE_viz} shows the ground truth and predictions for the traffic dataset, along with the mean of the absolute value of the residuals ($\mathbb{E}[|\hat{\mathbf{r}}|]$). It can be seen that for regions where the original mean of the predictions ($\mathbb{E}[\hat{\mathbf{y}} + \hat{\mathbf{r}}]$) are farther away from the ground truth, the standard deviation after EAE are also relatively larger. This indicates that $\mathbb{E}[|\hat{\mathbf{r}}|]$ can indeed capture the uncertainty of the predictions and therefore optimize the CRPS.

%%%%%%%%%%%%%%%%%%%%%%%%%%%%%%%%%%%%%%%%%%%%%%%%%%%%%%%%%%%%%%%%%%%%%%%%%%%%%%%%%%
%%%%%%%%%%%%%%%%%%%%%%%%%%%%%%%%%%%%%%%%%%%%%%%%%%%%%%%%%%%%%%%%%%%%%%%%%%%%%%%%%%
%%%%%%%%%%%%%%%%%%%%%%%%%%%%%%%%%%%%%%%%%%%%%%%%%%%%%%%%%%%%%%%%%%%%%%%%%%%%%%%%%
\section{Limitations and Future Work}
\label{sec:limits_and_future}

We point out several limitations and outline directions for future work. First, EAE currently works in theory only when predictions follow a Gaussian distribution; generalization to arbitrary predictive distributions is needed. Second, the PICP distance is computed using only three intervals; we aim to develop a more generalized metric that incorporates a broader range of intervals. Third, our results are not fully optimized with respect to hyperparameters; automatic hyperparameter tuning could further improve performance. Lastly, we assume that training errors follow a zero-mean Gaussian distribution, and future work can explore more flexible modeling of residuals.

%%%%%%%%%%%%%%%%%%%%%%%%%%%%%%%%%%%%%%%%%%%%%%%%%%%%%%%%%%%%%%%%%%%%%%%%%%%%%%%%%%
%%%%%%%%%%%%%%%%%%%%%%%%%%%%%%%%%%%%%%%%%%%%%%%%%%%%%%%%%%%%%%%%%%%%%%%%%%%%%%%%%%
%%%%%%%%%%%%%%%%%%%%%%%%%%%%%%%%%%%%%%%%%%%%%%%%%%%%%%%%%%%%%%%%%%%%%%%%%%%%%%%%%%

\ % The empty page

\end{document}